%% file: new_version.tex
\documentclass{article}

\usepackage[final]{neurips_2025}

\input{preamble}

\usepackage[utf8]{inputenc} 
\usepackage[T1]{fontenc}    
\usepackage{hyperref}       
\usepackage{url}            
\usepackage{booktabs}       
\usepackage{amsfonts}       
\usepackage{nicefrac}       
\usepackage{microtype}      
\usepackage{xcolor}         
\usepackage{graphicx} 
\usepackage{soul}
\usepackage{xcolor}
\definecolor{commentcolor}{HTML}{FFD7D7}

\usepackage{xr}
\usepackage{multirow}
\usepackage{makecell} 
\usepackage{pifont}
\newcommand{\cmark}{\ding{51}} 
\newcommand{\xmark}{\ding{55}} 
\usepackage{amsmath} 
\usepackage[most]{tcolorbox}
\definecolor{skyblue}{RGB}{100,180,255}
\usepackage{caption}
\title{Two Causally Related Needles in a Video Haystack}

\author{%
  Miaoyu Li\thanks{Equal contribution}, \,\, Qin Chao\footnotemark[1], \, and Boyang Li  \\
 College of Computing and Data Science \\
Nanyang Technological University, Singapore
\\
  \texttt{\{miaoyu.li, chao0009, boyang.li\}@ntu.edu.sg} \\
}

\begin{document}

\maketitle

\input{sec/0_abstract}

\input{sec/1_intro}
\input{sec/2_literature_review}
\input{sec/3_method}

\input{sec/4_experiments}
\input{sec/5.conclusion}
{
\bibliographystyle{plainnat}
\bibliography{main}
}

\newpage
\section*{NeurIPS Paper Checklist}

\begin{enumerate}

\item {\bf Claims}
    \item[] Question: Do the main claims made in the abstract and introduction accurately reflect the paper's contributions and scope?
    \item[] Answer: \answerYes{}{} 
    \item[] Justification: In the abstract, we clearly state our primary contribution, which is the design of a novel set of benchmark questions specifically created to evaluate models' abilities in long-context video understanding. In the introduction, we highlight two key limitations of existing video understanding datasets to underscore the necessity and effectiveness of our proposed benchmark. We further summarize the performance of contemporary video large language models and identify their strengths and weaknesses. Thus, these claims represent the primary contributions of our paper, and these contributions are also summarized explicitly in the introduction. 

\item {\bf Limitations}
    \item[] Question: Does the paper discuss the limitations of the work performed by the authors?
    \item[] Answer: \answerYes{} 
    \item[] Justification: In Appendix Section \ref{app:limits}. 

\item {\bf Theory assumptions and proofs}
    \item[] Question: For each theoretical result, does the paper provide the full set of assumptions and a complete (and correct) proof?
    \item[] Answer: \answerNA{} 
    \item[] Justification: This paper does not include theoretical results. 

    \item {\bf Experimental result reproducibility}
    \item[] Question: Does the paper fully disclose all the information needed to reproduce the main experimental results of the paper to the extent that it affects the main claims and/or conclusions of the paper (regardless of whether the code and data are provided or not)?
    \item[] Answer: \answerYes{} 
    \item[] Justification: For question generation, the procedure are introduced in Section~\ref{sec:data_create}. For model evaluation, We comprehensively introduced the models used in our experiments as well as the experimental procedures in Section \ref{sec:experiments}. Additionally, we have publicly released both the code and benchmark dataset to facilitate reproduction and verification of our main experimental results.

\item {\bf Open access to data and code}
    \item[] Question: Does the paper provide open access to the data and code, with sufficient instructions to faithfully reproduce the main experimental results, as described in supplemental material?
    \item[] Answer: \answerYes{} 
    \item[] Justification: Following the guidelines, we have fully disclosed all information necessary to reproduce our main experimental results. We have publicly released both our benchmark dataset (available at: \url{https://huggingface.co/datasets/causal2needles/Causal2Needles}) and the code used for evaluating model performance (available at: 
\url{https://github.com/jdsannchao/Causal2Needles}). For proprietary models, we set the \texttt{temperature} parameter to 0 to minimize randomness in generation. For open-source models, we set \texttt{do\_sample} to \texttt{False} to ensure deterministic outputs. These settings allow our reported results to be reliably reproduced. 

\item {\bf Experimental setting/details}
    \item[] Question: Does the paper specify all the training and test details (e.g., data splits, hyperparameters, how they were chosen, type of optimizer, etc.) necessary to understand the results?
    \item[] Answer: \answerYes{} 
    \item[] Justification: In Section \ref{sec:experiments}.

\item {\bf Experiment statistical significance}
    \item[] Question: Does the paper report error bars suitably and correctly defined or other appropriate information about the statistical significance of the experiments?
    \item[] Answer: \answerNA{} 
    \item[] Justification: To minimize randomness and ensure reproducibility, all models were evaluated using fixed decoding settings. Specifically, we set \texttt{temperature} to 0 for proprietary models and \texttt{do\_sample} to \texttt{False} for open-source models. As a result, model outputs are consistent across runs, and standard statistical significance tests or error bars are not applicable in this setting. 

\item {\bf Experiments compute resources}
    \item[] Question: For each experiment, does the paper provide sufficient information on the computer resources (type of compute workers, memory, time of execution) needed to reproduce the experiments?
    \item[] Answer: \answerYes{}{} 
    \item[] Justification: In Appendix Section \ref{app:GPUs}. 
    
\item {\bf Code of ethics}
    \item[] Question: Does the research conducted in the paper conform, in every respect, with the NeurIPS Code of Ethics \url{https://neurips.cc/public/EthicsGuidelines}?
    \item[] Answer: \answerYes{} 
    \item[] Justification: Yes, the research conducted in this paper fully conforms to the NeurIPS Code of Ethics. 

\item {\bf Broader impacts}
    \item[] Question: Does the paper discuss both potential positive societal impacts and negative societal impacts of the work performed?
    \item[] Answer: \answerNA{} 
    \item[] Justification: Our method primarily influences the evaluation of large video language models and may encourage the development of stronger video understanding capabilities. It does not have any direct societal impact, nor does it pose any foreseeable negative consequences.
    
\item {\bf Safeguards}
    \item[] Question: Does the paper describe safeguards that have been put in place for responsible release of data or models that have a high risk for misuse (e.g., pretrained language models, image generators, or scraped datasets)?
    \item[] Answer: \answerNA{} 
    \item[] Justification: No additional safeguards were required, as the dataset is used strictly for its intended academic purpose of evaluating model performance. The datasets used in our study are all publicly available, originally sourced from YouTube videos. We do not introduce any new high-risk data or models. Our benchmark is constructed purely for academic evaluation purposes and does not contain sensitive, private, or potentially misusable content. 

\item {\bf Licenses for existing assets}
    \item[] Question: Are the creators or original owners of assets (e.g., code, data, models), used in the paper, properly credited and are the license and terms of use explicitly mentioned and properly respected?
    \item[] Answer: \answerYes{} 
    \item[] Justification: All referenced datasets have been properly cited. 

\item {\bf New assets}
    \item[] Question: Are new assets introduced in the paper well documented and is the documentation provided alongside the assets?
    \item[] Answer: \answerYes{} 
    \item[] Justification: All new assets introduced in the paper, including our benchmark dataset and evaluation scripts, are well documented. 

\item {\bf Crowdsourcing and research with human subjects}
    \item[] Question: For crowdsourcing experiments and research with human subjects, does the paper include the full text of instructions given to participants and screenshots, if applicable, as well as details about compensation (if any)? 
    \item[] Answer: \answerNA{} 
    \item[] Justification: There is no crowd sourcing experiments.  

\item {\bf Institutional review board (IRB) approvals or equivalent for research with human subjects}
    \item[] Question: Does the paper describe potential risks incurred by study participants, whether such risks were disclosed to the subjects, and whether Institutional Review Board (IRB) approvals (or an equivalent approval/review based on the requirements of your country or institution) were obtained?
    \item[] Answer: \answerYes{} 
    \item[] Justification: This paper does not involve any potential risks. For human evaluation of the dataset, we obtained IRB approvals. 

\item {\bf Declaration of LLM usage}
    \item[] Question: Does the paper describe the usage of LLMs if it is an important, original, or non-standard component of the core methods in this research? Note that if the LLM is used only for writing, editing, or formatting purposes and does not impact the core methodology, scientific rigorousness, or originality of the research, declaration is not required.
    \item[] Answer: \answerYes{} 
    \item[] Justification:  The use of LLMs is a central component of our methodology. LLMs are involved in every step of our question generation pipeline. We explicitly describe the role of LLMs in Section \ref{sec:data_create}.

\end{enumerate}

\newpage
\appendix
\input{sec/9_appendix}


\end{document}

%% file: preamble.tex
\usepackage{amsthm}

%% file: sec/0_abstract.tex
\newcommand{\shortname}{\textsc{Causal2Needles}{}}

\begin{abstract}

Properly evaluating the ability of Video-Language Models (VLMs) to understand long videos remains a challenge. We propose a long-context video understanding benchmark, \shortname{}, that assesses two crucial abilities insufficiently addressed by existing benchmarks: (1) extracting information from two separate locations (two needles) in a long video and understanding them jointly, and (2) modeling the world in terms of cause and effect in human behaviors. \shortname{} evaluates these abilities using noncausal one-needle, causal one-needle, and causal two-needle questions. The most complex question type, causal two-needle questions, require extracting information from both the cause and effect events from a long video and the associated narration text. To prevent textual bias, we introduce two complementary question formats: locating the video clip containing the answer, and verbal description of a visual detail from that video clip. Our experiments reveal that models excelling on existing benchmarks struggle with causal 2-needle questions, and the model performance is negatively correlated with the distance between the two needles. These findings highlight critical limitations in current VLMs. The project page is available at: \href{https://limiaoyu.github.io/Causal2Needles/}{https://limiaoyu.github.io/Causal2Needles}
\end{abstract}

%% file: sec/1_intro.tex
\section{Introduction}
\label{sec:intro}

On many popular benchmarks \citep{mmlu, cobbe2021gsm8k, srivastava2022bigBench, chen2021evaluating_code}, recent AI systems achieved performance comparable to humans. However, real-world observations suggest that there is still a significant gap between AI and human capabilities \citep{west2023generative,frieder2023mathematical, amirizaniani2024llms, chinchure2025black}. This apparent contradiction suggests that (1) there is much overfitting to popular benchmarks such as GSM8K \citep{mirzadeh2024gsm} through mechanisms like data contamination \citep{singh2024evaluation}, and (2) existing benchmarks may not fully reflect differences between machine intelligence and human intelligence \citep{dziri2023faith,wu2024reasoning}. As a result, the development of sophisticated benchmarks that critically evaluate model capabilities has become a high priority for AI research. 

Within the context of long video understanding, we investigate two crucial limitations of existing benchmarks for Video-Language Models (VLMs). First, benchmarks evaluating information extraction from a single location do not fully reflect the long video understanding ability of VLMs. A popular type of evaluation for long-context models adopts the ``needle in a haystack'' problem formulation \citep{kamradt2023, wang2024needle}, where the ``needle'' represents the information to be extracted from the long context. Some long video benchmarks feature 1-needle questions that require information from a single location in the video. However, research in NLP indicates that model performances on 1-needle questions are often much higher than performances on questions requiring the extraction and understanding multiple needles, revealing the limitations of 1-needle questions for assessing long-context understanding \citep{li2023loogle, vodrahalli2024michelangelo, yang2024retrieval, levy2024same}. However, multiple-needle questions remain rare in the multimodal setting.

Second, existing benchmarks offer incomplete evaluation of whether VLMs possess an internal world model, which captures the underlying mechanisms governing object dynamics and human behaviors and enables predictions of future events from an intervention \citep{forrester1971, ha2018world}. Existing evaluations of world models \citep{guan2024world, motamed2025generative_physics_worldmodel, kang2024far_physics_worldmodel} focus solely on object motion prediction, neglecting human behaviors and event causality \citep{Sun2023EventCI}. 

To address these limitations, we propose a long-context video understanding benchmark, \shortname{}, comprising 2,606 1-needle questions and 1,494 2-needle questions. Out of these, 902 1-needle questions and all 2-needle questions involve causal, world-model reasoning. \shortname{} evaluates two key abilities of VLMs: (1) extracting relevant information from two locations in long videos and jointly reasoning about them, and (2) modeling the world in terms of causes and effects of human behaviors. 

\begin{figure*}[t] 
    \centering
    \includegraphics[width=0.93\linewidth]{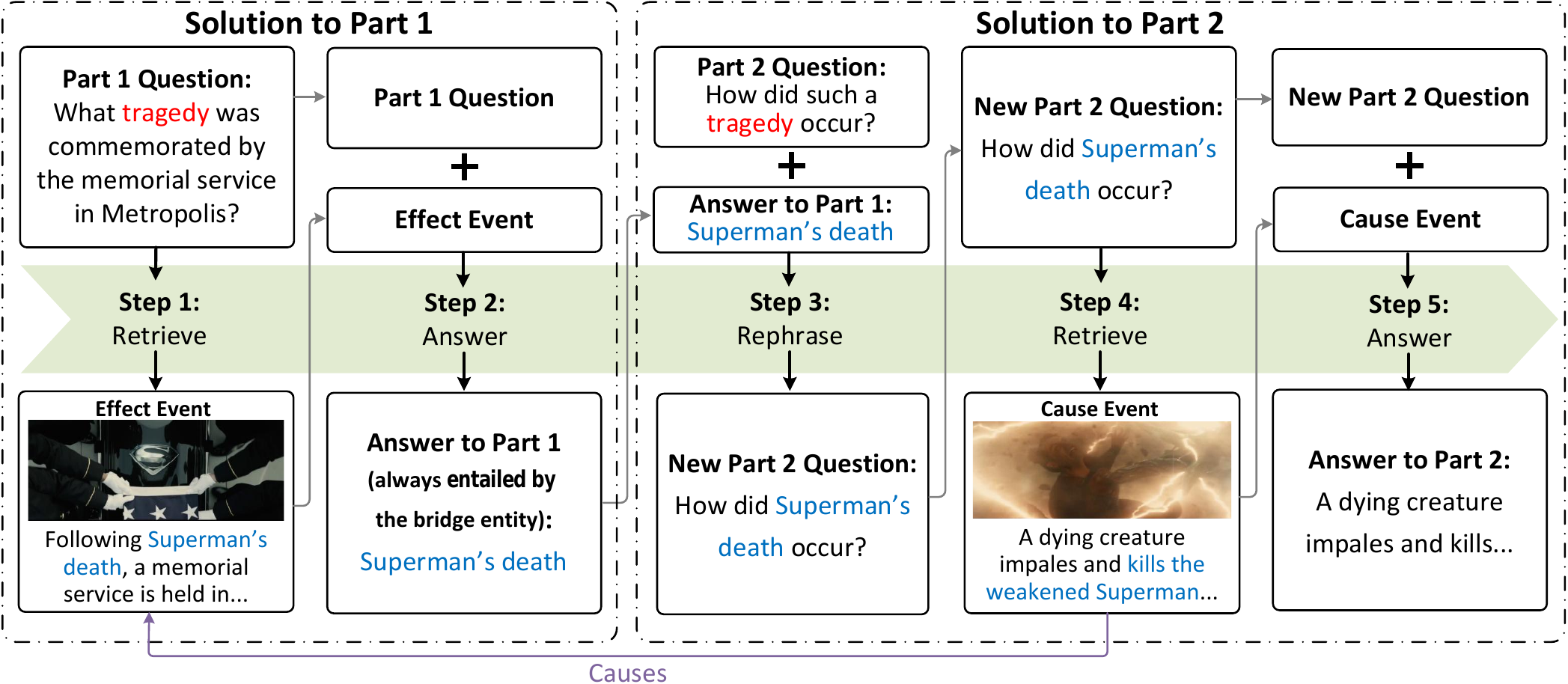} 
    \caption{The logical solution process for the 2-needle questions of \shortname. Each step involves an operation, with the input shown above the step and the output below the step. The question purposely refer to the bridge entity, ``Superman's death,'' ambiguously as ``tragedy.'' As a result, one must first resolve the bridge entity using Part 1 before answering Part 2. This question design mandates joint understanding of both the cause and effect events. Note that the steps are necessary only in an information-processing sense. A VLM may adopt different steps. }
    \label{fig:bridge entity}
    \vspace{-0.3cm}
\end{figure*}

The causal 2-needle questions are constructed from a pair of cause and effect events. Each question should require the VLM to first retrieve the effect event and then the cause event. To formulate the question, we identify a \emph{bridge entity}, which is an entity or a piece of information shared by the cause event and the effect event. Part one of the question asks the VLM to resolve the bridge entity by retrieving the effect event. Part two of the question requires the retrieval of the cause of the effect event. As an example, Fig. \ref{fig:bridge entity} shows a cause event and an effect event that share a bridge entity, ``Superman's death,'' ambiguously referred to as ``tragedy.'' To answer question part one, the model must resolve the content of the tragedy by retrieving the effect event, which reveals the tragedy is Superman's death. Only after that, the model can answer question part two by retrieving the video clip showing the cause of Superman's death. If the bridge entity were explicitly stated, the question would become ``how did Superman die, leading to his memorial service in Metropolis'', which can be answered by retrieving the cause event directly. In that case, the 2-needle question would degenerate to a 1-needle question. This novel problem formulation allows \shortname{} to assess the two-needle ability and the causal reasoning ability together.   


\begin{figure*}[h] 
    \centering
    \includegraphics[width=\linewidth]{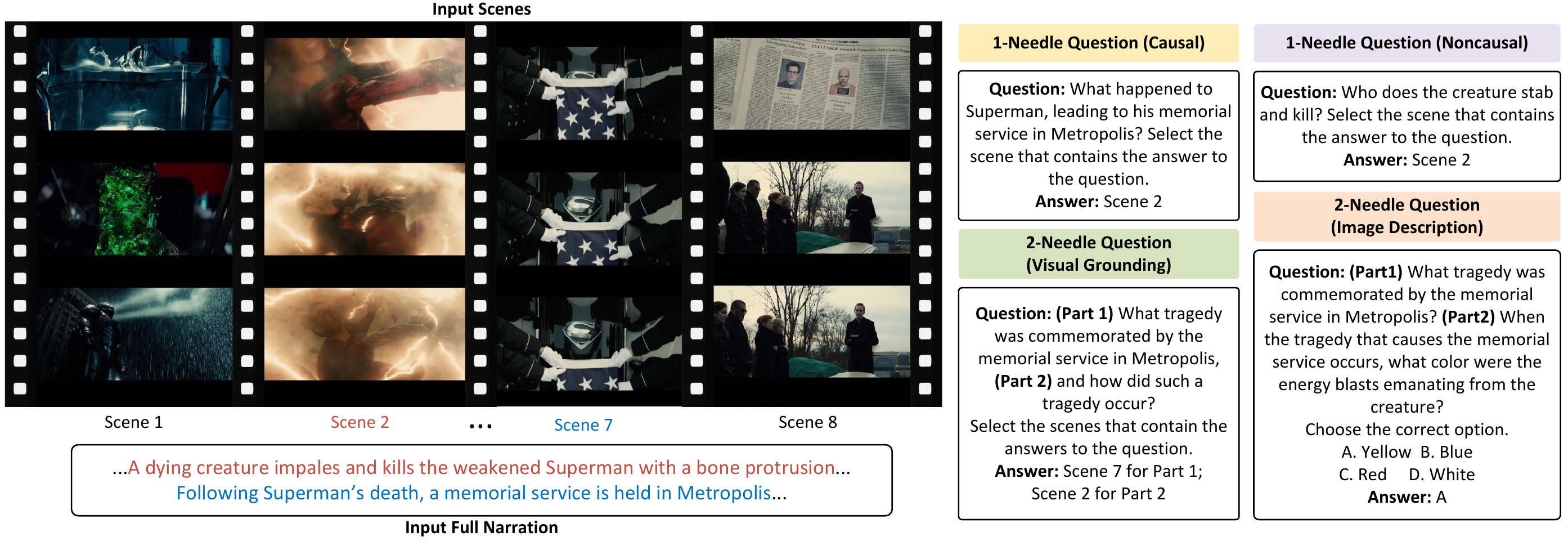} 
    \caption{The evaluation framework of \shortname{}. To help models understand the storyline, we also feed the full textual narration into the model. Four types of questions are designed for each pair of causally related events.}
    \label{fig:framework}
    \vspace{-0.3cm}
\end{figure*}
\shortname{} is constructed on top of movie summary videos, containing video clips from the movies and narration text. However, this may create a shortcut that VLMs can exploit. The video clips could be difficult to understand without the narration text, but if we feed the narration text as input, the VLM may answer the question from the text directly without using the video. This is the infamous  phenomenon of textual bias  \citep{ko2023_llm_are_reasoner_for_video, cores2024tvbench, xiao2024can}. 

As mitigation, we introduce two complementary question formats. The first format, called \emph{visual grounding}, requires the model to select the video clip corresponding to the event it needs to retrieve. The second format, called \emph{image description}, requires the model to answer multiple-choice questions about the appearance of the retrieved video clip. Fig.~\ref{fig:framework} shows some examples. Visual grounding questions necessitate the understanding of video clips, but they may underestimate model performance as the format may be out-of-distribution (OOD) to most VLMs. In contrast, image description questions circumvent the OOD issue, but the VLM may benefit from knowledge of the movie learned from pretraining data, such as the color of Lois Lane's outfit, leading to overestimation of model abilities. \shortname{}  adopts the visual grounding format for all 1-needle questions and splits the 2-needle questions between the two formats (747 each).
Experiments reveal several important findings. First, causal questions appear substantially more difficult than noncausal questions. Second, models that perform well on 1-needle questions exhibit remarkable performance drops on 2-needle questions. Finally, the distance between the two needles is negatively correlated with model performance. Taken together, these results demonstrate that joint understanding of and causal reasoning over two separate video locations remain important weaknesses of existing VLMs. We summarize the contributions of this paper as follows:




$\bullet$ We propose the \shortname{} benchmark that contains causal 1-needle and 2-needle questions, generated from story videos. We devise a question creation strategy utilizing bridge entities to force the video-language models to jointly understand two video clips arbitrarily located in the context window. 

$\bullet$  We reveal significant weaknesses in current VLMs in the joint understanding and retrieval of two separate video clips, even though their 1-needle performances are high. The more distant the two clips are, the worse the models perform. 






%% file: sec/2_literature_review.tex
\section{Related Work}





\subsection{Long Video Understanding Benchmarks}

Retrieval of information from a specific location in a long video (usually called ``needle in a haystack'') is widely regarded as a key capability in long-video understanding  \citep{kamradt2023, wang2024needle}. In Tab.~\ref{tab:literature_review}, we compare recent long-video benchmarks with multiple-needles-in-haystack tasks side by side to highlight the unique contribution of the proposed \shortname{} dataset.
%
%
EgoSchema \citep{mangalam2023egoschema} does not measure specific model capabilities (\textit{e.g.}, 1-needle vs. 2-needle, reasoning vs. recognition) as the questions are not categorized by the required capability, resulting in limited diagnostic precision.
MLVU \citep{zhou2024mlvu} require models to retrieve visually distinct images that are artificially inserted into videos, which may allow for  shortcut features due to the domain gap between the artificial needle and the original content. 
Multi-needle questions also exist in MVBench \citep{li2024mvbench} and TVBench \citep{cores2024tvbench}, which contain questions to evaluate the model ability to track an object across multiple temporal locations. 
Similar questions are also featured in VideoMME \citep{fu2024videomme} (\textit{e.g.}, counting problems)  and LongVideoBench (\textit{e.g.}, L2-SSS questions) \citep{wu2025longvideobench}. However, questions in the above datasets typically require only independent understanding of each needle. Moreover, the multiple needles often only related by  surface-level visual cues, such as matching visual appearances of objects, rather than semantic relations like causality.

\input{tables/related_works}


In contrast, the \shortname{} benchmark requires models to jointly understand both the cause and the effect---the understanding of the cause relies on the correct interpretation of the clue in the effect. Furthermore, the two needles are related by causality rather than visual similarity, necessitating semantic understanding. These designs establish a challenging and unique benchmark for long-video understanding.

\subsection{World Model}

There is an ongoing debate \citep{gurnee2023language_worldmodel, liu2024world, guan2024worldmodel_drive, hu2024_math_case_worldmodel, kang2024far_physics_worldmodel, motamed2025generative_physics_worldmodel} over whether deep neural networks learn to develop internal world models---whether they can identify the underlying principles, such as  laws of physics, that govern the observational data they learn from. Such world models, if present, can be used to predict future outcomes of possible interventions, such as kicking a ball or blowing a candle \citep{forrester1971, ha2018world, FLASH-2025}. 

However, this debate largely overlooks causality between events involving humans, which could be induced by factors like biology (e.g., heavy sweating causes thirst), psychology (e.g., hearing compliments makes people feel good), and social norms (e.g., providing good service leads to tipping). Although causal questions sporadically appear in video understanding datasets, such as the episodic reasoning questions in MVBench, it is usually difficult to isolate them and quantitatively measure this factor independently.  In contrast, \shortname{} is dedicated to evaluating reasoning over cause and effect in the context of human behaviors, which allows precise diagnosis of this capability and fills a gap in benchmarks of world models.




%% file: tables/related_works.tex
\begin{table}
\centering

\caption{A comparison of \shortname{} with other multi-needle long video benchmarks. \shortname{} is the only benchmark dedicated to needle-in-haystack problems (Diagnostic Precision) that requires joint understanding of the two needles and the identification of cause events from effect events.  }\label{tab:literature_review}
\resizebox{\textwidth}{!}{\begin{tabular}{@{}llrccccc@{}}
\toprule
     Benchmark & Video Length  &  \# QA  & \makecell[c]{Diagnostic \\ Precision}  & \makecell[c]{Needle \\Type} & \makecell[c]{Joint \\Understanding} & \makecell[c]{Causal \\ Reasoning}   \\
\midrule
     EgoScheme \citep{mangalam2023egoschema} &         180 s &   5,000 &                     \xmark &                  Natural &     \xmark&                \cmark  \\
       MVBench \citep{li2024mvbench} &     16 s-40 s &   4,000 &                     \cmark &                  Natural &     \xmark&                \cmark \\
       TVBench \citep{cores2024tvbench} &     16 s-40 s &   2,525 &                     \cmark &                  Natural &     \xmark&                \xmark  \\
          MLVU \citep{zhou2024mlvu} &    180 s-3600 s &   3,102 &                     \cmark &               Artificial &     \xmark&                \xmark  \\
      VideoMME \citep{fu2024videomme} &        1018s  &   2,700 &                     \cmark &                  Natural &     \xmark&                \cmark  \\
LVB \cite{wang2024lvbench} &        473 s  &   6,679 &                     \cmark &                  Natural &     \xmark&                \xmark  \\
\shortname{} (Ours)  &        438 s  &   4,100 &                     \cmark &                  Natural &     \cmark&                \cmark \\
\bottomrule
\end{tabular}
}
\vspace{-0.3cm}
\end{table}

%% file: sec/3_method.tex
\section{\shortname{}: Dataset Construction}
\label{sec:data_create}

\shortname{} is built on two video-language datasets, YMS \citep{yms} and SyMoN \citep{symon}. With a total of 192 fully annotated movie recap videos, the datasets offer a rich collection of human-behavior events from diverse movie genres. Each event consists of a narration sentence and its corresponding video clip.
Details such as the distribution of movie themes and the temporal distance between causal events can be found in Appendix Sec.~\ref{app:stat}.


To facilitate the evaluation of models of diverse capabilities,  \shortname{} consists of both 1-needle and 2-needle questions, as well as causal and noncausal questions, which have different difficulty levels. 
The generation of both 1-needle and 2-needle questions depend on  pairs of cause and effect sentences, extracted in a causal relationship extraction step (Sec. \ref{method:extraction}). 
We describe the generation of 1-needle questions in Sec. \ref{method:1n-gen} and the generation of 2-needle questions in Sec. \ref{method:2n-gen}.

\subsection{Causal Relationships Extraction}
\label{method:extraction}
We employ a Large Language Model (LLM) to extract causal relationships from narration by combining global and local event graphs, based on an event graph extraction method \citep{Sun2023EventCI}. Each event (a narration sentence) is represented as a node, and causal relationships as directed edges. The global graph is extracted from the complete narration of a video. The graph captures long-range causal relationships but may be incomplete, because LLM often overlook sentences in the middle section of a long context \citep{liu-etal-2024-lost}. To address this, we introduce a sliding-window approach that extracts local graphs from 15-sentence segments with a 5-sentence stride. These local graphs capture more detailed causal relationships with a shorter range. We merge the global and local graphs to obtain comprehensive and long-range causal relationships. To avoid superficial causal relationships resulting from temporal adjacency, we retain only those where the cause and effect are separated by at least three events. In \shortname{}, the distance between cause and effect events ranges from 3 to 21 events. Appendix Sec. \ref{app:stat} and \ref{app: merge} contain more details. The prompt used is shown in Appendix Fig. \ref{fig: causal_extraction}.

\subsection{Generation of One-Needle Questions}
\label{method:1n-gen}
With the extracted causal relationships, we use an LLM to generate simple 1-needle questions. To facilitate reader understanding, we illustrate the process using the cause and effect events shown in Fig. \ref{fig:framework}. 
We first prompt the LLM to generate noncausal one-needle questions, which ask for a detail in either the cause or the effect event. For the cause event in Fig. \ref{fig:framework}, the generated question is: ``Who does the creature stab and kill?'' We can similarly generate another question for the effect event.
The causal one-needle questions differ from the noncausal question by requiring understanding of the causal relation between the two events. We prompt the LLM to generate a question asking for the cause event of a specific effect. The generated question for the example in Fig. \ref{fig:framework} is: ``What happened to Superman, leading to his memorial service in Metropolis?''
%
%
Finally, to mitigate potential textual bias from the input narration, we combine these questions with the visual grounding instruction: ``Select the scene that contains the answer to the question.'' The prompt for 1-needle question generation is shown in Appendix Fig. \ref{fig:cau-1-needle}.

\subsection{Generation of Two-Needle Questions}
\label{method:2n-gen}

Generating 2-needle questions involves two steps after obtaining causal relationships: (1) rephrasing the cause and effect sentences to establish the bridge entity between them, and (2) generating a question that requires joint understanding of both cause and effect events.

\paragraph{Rephrasing the Cause and Effect Sentences.}
The generation of 2-needle questions is based on a bridge entity that connects cause and effect events. The bridge entity serves two critical purposes: it provides a hint that drives the model to locate the cause event, so we do not provide the cause event directly; it forces the model to  resolve the content of the vaguely phrased bridge entity by locating the effect event, before it can locate the cause event. 

However, the bridge entity may not be explicitly mentioned in the original cause and effect sentences. Therefore, we prompt a VLM to rephrase the cause and effect sentences to explicitly establish the bridge entity. For example, the cause sentence, ``As it dies, the creature stabs and kills the weakened Superman with one of its bone protrusions,'' and the effect sentence, ``A memorial is held for Superman in Metropolis,'' are rephrased into sentences in Fig. \ref{fig:framework}. To maintain semantic consistency between the rephrased sentences and their corresponding clips, we also input the video clips into the VLM as a constraint. 

\paragraph{Generation of Visual Grounding Questions.}
After obtaining causal sentence pairs with a clear bridge entity, we utilize an LLM to generate two parts of a visual grounding 2-needle question. We use the causal relationship in Fig. \ref{fig:framework} to demonstrate this process.
%
First, we prompt the LLM to generate a Part 1 question that requires retrieving the effect event to identify the bridge entity. Specifically, we ask it to extract the bridge entity that establishes the causal relationship, such as ``Superman's death.'' Next, we instruct the LLM to rephrase this bridge entity as a vague reference, such as ``tragedy.'' After that, using this vague reference, we prompt the LLM to create a question that uses the effect sentence as context and the bridge entity as the answer. In this case, the final output is: ``What tragedy was commemorated by the memorial service in Metropolis?''
Further, we prompt the LLM to generate a Part 2 question, which requires retrieving the cause event. In the example, with the vague reference ``tragedy'', we instruct the LLM to generate a question whose answer appears exclusively in the cause sentence. The output is: ``How did such a tragedy occur?'' It is important to keep the bridge entity vague in this question, so that the question does not give the effect event away. Instead, the model must retrieve the video clip of the effect event and resolve the bridge entity reference.
%
%
Finally, combining the two question parts with a task instruction, we complete a visual grounding 2-needle question: ``What tragedy was commemorated by the memorial service in Metropolis, and how did such a tragedy occur? Select the scenes that contain the answers to the question.'' To answer it, the model must jointly understand cause and effect events and ground the answers on video clips. The prompts are in Appendix Fig. \ref{fig:cau-2-needle}.

\paragraph{Generation of Image Description Questions.}
The visual grounding question format may be out-of-distribution for some models. As a result, it may underestimate their performance. To fix this, we also generate questions in a complementary format: multiple-choice image description questions. These questions share the same Part 1 as visual grounding questions, but modify Part 2 to ask about visual details of the cause event.

With the cause video clip, the cause sentence, and the effect sentence as input, we prompt an LLM to generate a question that asks for an attribute of a visible object in the clip. This question follows the template: ``When the event that causes $\langle \texttt{Effect Event} \rangle$ occurs, $\langle \texttt{Image Description Question}\rangle$.'' For example, in the clip where Metropolis mourns Superman, 
the LLM output could be: ``When the battle that causes Superman’s death and leads to the memorial service occurs, what color were the energy blasts emanating from Doomsday?'' We then further prompt the LLM to generate four challenging answer options based on the question.

However, the generated questions may reveal the bridge entity or include excessive object details, allowing the model to locate the cause clip without truly understanding the Part 1 question. For instance, the initially generated question reveals the bridge entity ``Superman's death'' and the name ``Doomsday'', which reduces the difficulty of locating the video clip. To mitigate this issue, we prompt the LLM to obscure the bridge entity and the named entity being inquired about. 
We then obtain the final image description 2-needle question, which begins with the Part 1 question: ``What tragedy was commemorated by the memorial service in Metropolis? When the tragedy that causes the memorial service occurs, what color were the energy blasts emanating from the creature? A. Yellow B. Blue C. Red D. White'' The prompts are shown in Appendix Fig. \ref{fig:vis-2-needle} and Fig. \ref{fig:vis-2-needle-options}.

\subsection{Quality Evaluation of Generated Questions} 

We conduct automatic evaluation and human evaluation of the quality of questions and bridge entities, which provide vital information for 2-needle questions. We evaluate 4 quality factors. Factor 1 is if the bridge entity is truly shared by the cause and effect events it is supposed to connect. We report the proportion of affirmative answers as the final score. Factor 2 is the correctness of purposely vague references to the bridge entities, or if the vague reference is indeed more ambiguous than the original bridge entity but still preserves its core meaning. The evaluation result for each vague reference should be ``Yes'' or ``No''. We report the proportion of ``Yes''. Factor 3 is the factual correctness of questions, or the extent to which a question is consistent with the story and does not introduce hallucination or contradiction to the story. We use a 5-point scale, where 1 is the lowest score and 5 the highest. Factor 4 is the readability of questions, as reflected by the naturalness, grammar, and clarity, on a 5-point scale.

To verify that the evaluation LLMs are not biased to always indicate high quality, we also create several random baselines. For Factor 1, we randomly match the bridge entities and the cause-effect event pairs, which should cause the LLM to answer No to shared existence of the bridge entity. For Factor 2, we randomly shuffle the correspondence between the bridge entities and the vague references. For Factor 3, we randomly shuffle the correspondence between the questions and the cause-effect event pairs. For Factor 4, due to the difficulty in writing unreadable questions, we do not construct any random baseline for readability.

We utilize two state-of-the-art models as the evaluation LLMs, ChatGPT-4.1 and Gemini-2.0-flash (neither is involved in the question generation process), and recruit five human annotators. The models evaluate both 1-needle and 2-needle questions, while the annotators evaluate 136 visual grounding 2-needle questions. For human evaluation, we adopt majority voting among the five annotators for Yes/No evaluations, and use the average for numerical evaluations. 

Tab. \ref{tab:question_quality} shows the evaluation results. The generated questions receive near-perfect scores on all metrics. Reassuringly, all random baselines score near the lowest possible, which is  0\% for shared existence of bridge entities and correctness of vague references, and 1 for factual correctness of questions. These results indicate that the automatically generated questions have high quality and the evaluation processes are valid. The prompts for question evaluation are in Appendix Sec. \ref{app:prompts}. 


\begin{table}[t]
  \centering
  \caption{Evaluation results of generated questions. VG and ID refer to visual grounding and image description, respectively. 1-N and 2-N denote 1-needle and 2-needle questions. Numbers in parentheses indicate the performance of random baselines.}
  \resizebox{\linewidth}{!}{
\begin{tabular}{@{}c c c cccc cccc@{}}
\toprule
\multirow{2}{*}{\textbf{Models}} 
  & \multirow{2}{*}{\makecell{\textbf{Shared Existence}\\\textbf{of Bridge Entities}}} 
  & \multirow{2}{*}{\makecell{\textbf{Correctness of}\\\textbf{Vague References}}} 
  & \multicolumn{4}{c}{\textbf{Factual Correctness of Questions}} 
  & \multicolumn{4}{c}{\textbf{Readability of Questions}} \\ 
\cmidrule(l){4-11} 
 & & 
 & \makecell{\textbf{Noncausal}\\\textbf{1-N}} 
 & \makecell{\textbf{Causal}\\\textbf{1-N}} 
 & \textbf{VG 2-N} 
 & \textbf{ID 2-N} 
 & \makecell{\textbf{Noncausal}\\\textbf{1-N}} 
 & \makecell{\textbf{Causal}\\\textbf{1-N}} 
 & \textbf{VG 2-N} 
 & \textbf{ID 2-N} \\ 
\midrule
ChatGPT-4.1 & 95.6\% (0.3\%) & 91.0\% (3.6\%) & 4.71 (1.10) & 4.62 (1.12) & 4.99 (1.10) & 4.74 (1.13) & 4.91 & 4.85 & 4.83 & 4.67 \\
Gemini-2.0-flash & 95.0\% (3.8\%) & 98.7\% (2.5\%) & 4.75 (1.05) & 4.66 (1.02) & 4.96 (1.01) & 4.83 (1.00) & 4.75 & 4.18 & 4.69 & 4.25 \\
\midrule
Human & 82.4\% & 98.5\% & - & - & 4.50 & - & - & - & 4.80 & - \\
\bottomrule
\end{tabular}

}
  \label{tab:question_quality}
\end{table}

%% file: sec/4_experiments.tex
\begin{table*}[t]
  \centering
      \caption{Quantitative results (accuracy, \%) of VLMs on our benchmark. ``Forward'' refers to inputting video clips in chronological order, while ``Reverse'' uses reverse order. ``Avg'' denotes results averaged over both orders. Best scores are in \textbf{bold}.}
    \resizebox{\linewidth}{!}{
\begin{tabular}{@{}ccccccccccccc@{}}
\toprule
\multirow{3}{*}{\textbf{Models}} &
\multirow{3}{*}{\begin{tabular}[c]{@{}c@{}}\textbf{Noncausal 1-N}\\ \textbf{Questions}\end{tabular}} &
\multirow{3}{*}{\begin{tabular}[c]{@{}c@{}}\textbf{Causal 1-N}\\ \textbf{Questions}\end{tabular}} &
\multicolumn{9}{c}{\textbf{VG 2-N Questions}} &
\multirow{3}{*}{\begin{tabular}[c]{@{}c@{}}\textbf{ID 2-N}\\ \textbf{Questions}\end{tabular}} \\ 
\cline{4-12}
& & &
\multicolumn{3}{c}{Forward} &
\multicolumn{3}{c}{Reverse} &
\multicolumn{3}{c}{Avg} &
\\ \cline{4-12}
& & &
Part 1 & Part 2 & \multicolumn{1}{c}{Both} &
Part 1 & Part 2 & \multicolumn{1}{c}{Both} &
Part 1 & Part 2 & Both &
\\ \toprule
Human & -- & 78.2 & 83.7 & 85.9 & \multicolumn{1}{c}{79.3} & \multicolumn{3}{c}{ - } & \multicolumn{3}{c}{ - } & 88.2 \\
\midrule
\multicolumn{1}{c}{\textit{Proprietary Models}} & \\
ChatGPT-4o           & \textbf{56.8} & \textbf{39.2} & 16.7 & 39.2 & \multicolumn{1}{c}{9.4} & \textbf{45.4} & 21.2 & \multicolumn{1}{c}{\textbf{13.4}} & \textbf{31.1} & 30.2 & \textbf{11.4} & 59.2 \\
Gemini-1.5-pro        & 55.4 & 35.6 & \textbf{21.0} & \textbf{40.0} & \multicolumn{1}{c}{\textbf{10.2}} & 35.7 & \textbf{21.4} & \multicolumn{1}{c}{8.4} & 28.4 & \textbf{30.7} & 9.3 & \textbf{60.9} \\
ChatGPT-4o-mini       & 39.9 & 33.4 & 17.4 & 22.9 & \multicolumn{1}{c}{5.0} & 32.4 & 11.9 & \multicolumn{1}{c}{5.2} & 24.9 & 17.4 & 5.1 & 52.3 \\
Claude-3.5-sonnet     & 37.6 & 26.5 & 16.6 & 22.4 & \multicolumn{1}{c}{4.8} & 19.3 & 13.9 & \multicolumn{1}{c}{2.9} & 17.9 & 18.1 & 3.9 & 60.5 \\
\midrule
\multicolumn{1}{c}{\textit{Open-source Models}} \\
Qwen2.5VL-32B          & \textbf{30.7} & 11.7  & 26.3 & 17.7 & \multicolumn{1}{c}{5.4} & 10.3 & 20.4 & \multicolumn{1}{c}{1.9} & 18.3 & 19.0 & 3.6 & \textbf{53.5} \\
Qwen2.5VL-7B          & 17.5 & 13.6 & \textbf{27.6} & \textbf{17.7} & \textbf{5.0} & 11.2 & \textbf{18.9} & \textbf{1.9} & \textbf{19.4} & \textbf{18.3} & \textbf{3.4} & 43.2 \\
LLaVA-Next-Video-34B  & 12.4 & 12.3 & 0.8 & 17.4 & \multicolumn{1}{c}{0.0} & 11.8 & 0.9 & \multicolumn{1}{c}{0.0} & 6.3 & 9.2 & 0.0 & 48.6 \\
LLaVA-OneVision-7B    & 12.3 & \textbf{18.0} & 4.6 & 14.7 & \multicolumn{1}{c}{0.0} & 17.0 & 5.6 & \multicolumn{1}{c}{0.1} & 10.8 & 10.2 & 0.1 & 28.3 \\
InternVL2-8B          & 11.6 & 7.4  & 14.5 & 8.3 & \multicolumn{1}{c}{1.2} & 9.5 & 9.1 & \multicolumn{1}{c}{0.5} & 12.0 & 8.7 & 0.9 & 40.2 \\
LLaVA-Next-Video-7B   & 11.7 & 17.2 & 0.0 & 4.4 & \multicolumn{1}{c}{0.0} & 15.9 & 0.0 & \multicolumn{1}{c}{0.0} & 8.0 & 2.2 & 0.0 & 27.3 \\
LongVA-7B             & 9.2 & 14.7 & 2.8 & 5.0 & \multicolumn{1}{c}{0.0} & 10.3 & 0.7 & \multicolumn{1}{c}{0.0} & 6.6 & 2.8 & 0.0 & 49.7 \\
Aria-28B              & 7.0 & 12.1 & 19.0 & 14.8 & \multicolumn{1}{c}{0.6} & \textbf{18.7} & 18.1 & \multicolumn{1}{c}{0.1} & 18.9 & 16.5 & 0.4 & 43.0 \\
LongVU-7B             & 3.3 & 12.2 & 3.2 & 1.3 & \multicolumn{1}{c}{0.3} & 4.4 & 2.1 & \multicolumn{1}{c}{0.5} & 3.8 & 1.7 & 0.4 & 34.2 \\
\midrule
Random Chance         & 9.8 & 9.8 & 9.8 & 9.8 & \multicolumn{1}{c}{1.0} & 9.8 & 9.8 & \multicolumn{1}{c}{1.0} & 9.8 & 9.8 & 1.0 & 25.0 \\
\bottomrule
\end{tabular}
}
  \label{tab:main_results}
\end{table*}

\section{Evaluation of VLMs}
\label{sec:experiments}
\subsection{Evaluation Setup}

We evaluate a total of 13 advanced VLMs, consisting of 4 proprietary models and 9 open-source models. Proprietary models include ChatGPT-4o (gpt-4o-2024-08-06) \citep{openai2024chatgpt4o}, ChatGPT-4o-mini (gpt-4o-mini-2024-07-18) \citep{openai2024chatgpt4o_mini}, Gemini-1.5-pro (gemini-1.5-pro-002, Sep 2024) \citep{team2024gemini}, and Claude-3.5-sonnet (claude-3-5-sonnet-20241022) \citep{anthropic2024claude3}. Open-source models include LLaVA-Next-Video-7B \citep{zhang2024video_llavanext}, LLaVA-Next-Video-34B \citep{zhang2024video_llavanext}, LLaVA-OneVision-7B \citep{li2024llava_onevision}, LongVU-7B \citep{shen2024longvu}, Aria-28B \citep{li2024aria}, Qwen2.5VL-7B/32B \citep{bai2025qwen2}, LongVA-7B \citep{zhang2024longva} and InternVL2-8B \citep{chen2024far_internvl}. To establish a human baseline, we employ two annotators to answer 316 randomly selected questions (including causal 1-needle, 2-needle visual grounding and image description). Details about the input content, the chat template of VLMs can be found in Appendix Sec.~\ref{app:GPUs}.

Most models we tested do not natively support inputs containing multiple video clips. As a workaround, we uniformly sample five frames from each clip and stack them vertically as one input image. This allows us to stay within the image input limit, which is 32 images for most models. Compared to other possible techniques we experimented with, this trick yields the best performance (see Appendix Sec. \ref{app:CE_length}).


To prevent the models from using sentence indices as shortcuts, we input the full narration along with only a subset of the corresponding video clips. The narration and video clips are provided separately. We avoid aligning video clips with sentences in the input to prevent the models from relying on text to locate video clips. We input all video clips from the cause clip to the effect clip, as well as a total of five clip before and after this span; the numbers of clip before and after are random. This setup prevents the cause clip from always being the first clip and the effect clip always the last, eliminating 
shortcut learning based on location.


\subsection{Main Results}
\label{sec:main_results}

We present the performance of various VLMs on the noncausal 1-needle questions, causal 1-needle questions, Visual Grounding (VG) 2-needle questions, and Image Description (ID) 2-needle questions in Tab. \ref{tab:main_results}. We report the accuracy on each type of questions. For VG 2-needle questions, we separately compute the accuracy for Part 1, Part 2, and both parts answered correctly.  In addition, we evaluate each model using both the original (forward) and reversed video clip order, and report the average of the two orders as the final result on VG 2-needle questions. This design is motivated by positional bias of the models, discussed in Sec.~\ref{sec:2bias}. 

\begin{figure}[t]
    \centering
    \includegraphics[width=0.7\textwidth]{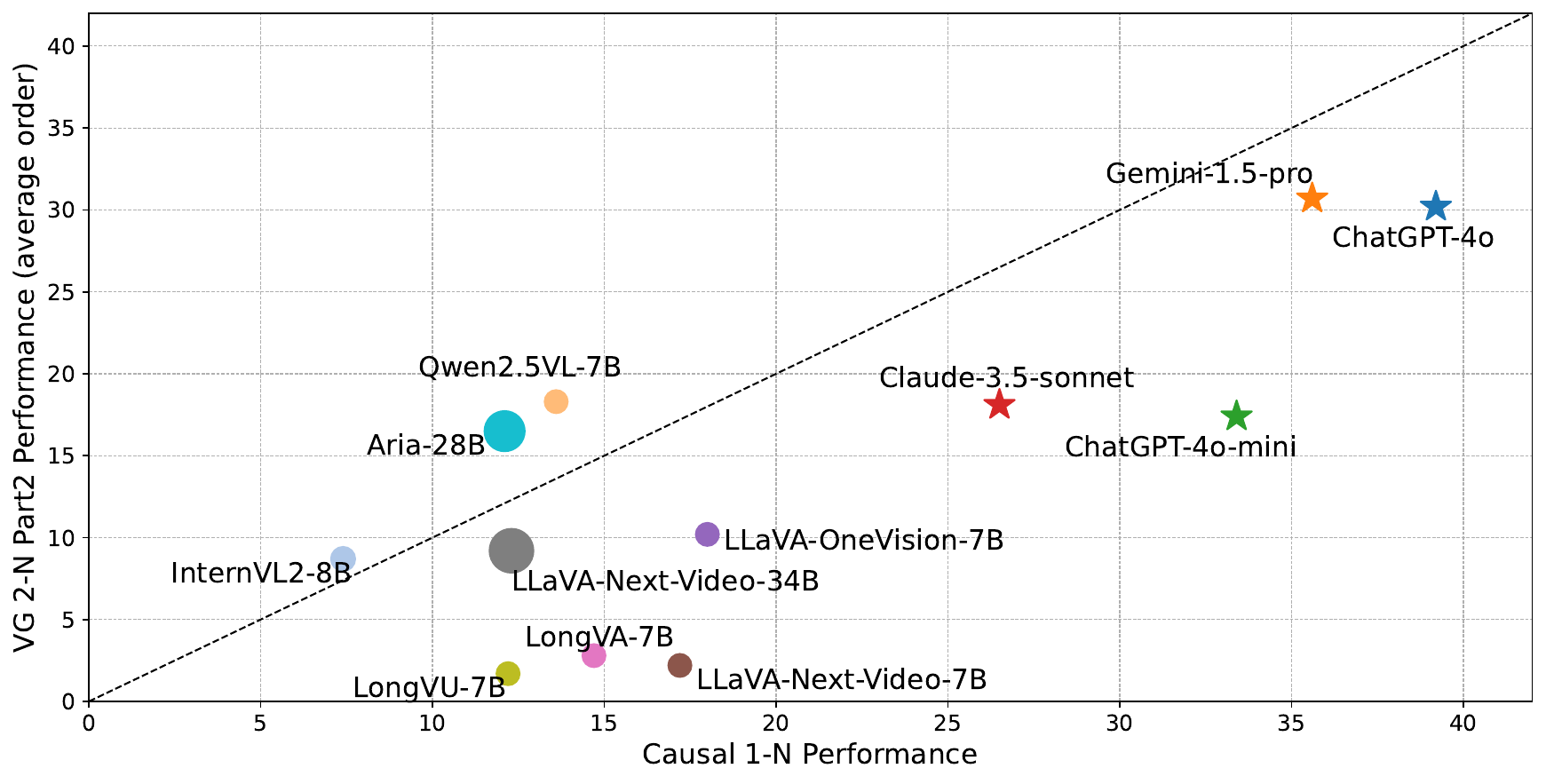}
    \caption{Comparison of  performance of models on Part 2 of VG 2-needle questions (average order) and causal 1-needle questions. This is a fair comparison since both ask to retrieve the cause event. The dashed diagonal line represents equal performance across the two question types. Most points fall below the line, indicating that models are poorer at VG 2-needle questions than causal 1-needle questions. The size of the dots indicates the model size. The stars indicate proprietary models.}
    \label{fig:1n_vs_2n}
\end{figure}

\begin{figure}[t]
    \centering
    \includegraphics[width=0.8\textwidth]{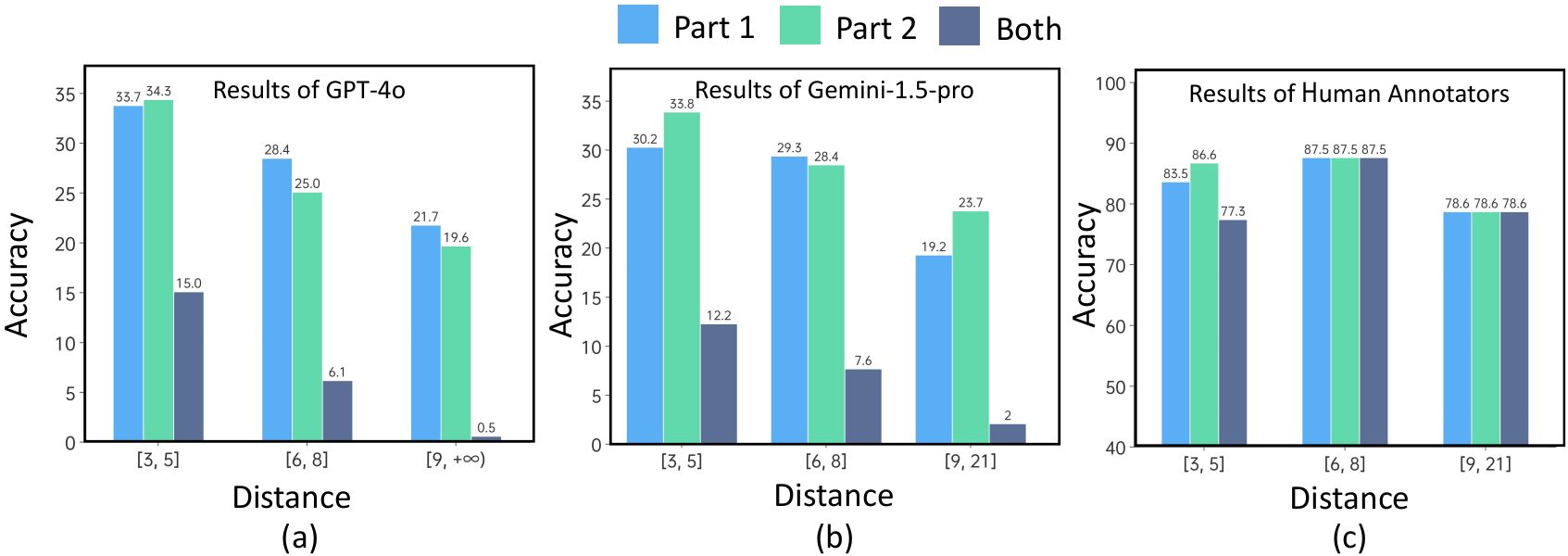}
    \caption{Performance on VG 2-needle questions as the distance between the two needle grows. We report the average-order performance for models and the forward-order performance for human annotators. The model performance declines whereas human performance stays mostly unchanged.}
    \label{fig:per_distance}
\end{figure}

\paragraph{Causal Questions Are More Challenging Than Noncausal Questions.}
From Tab. \ref{tab:main_results}, we observe that most models perform substantially better on noncausal 1-needle questions than on causal ones. For instance, ChatGPT-4o and Qwen2.5-VL-32B achieve results that are 17.6\% and 19.0\% higher, respectively. This gap suggests that causal reasoning still poses a significant hurdle for current state-of-the-art models on long-video understanding.

\paragraph{Two-needle Questions Are More Challenging Than One-needle.}

We compare the performance on
Part 2 of VG 2-needle questions (average order) and causal 1-needle questions, since both ask for the retrieval of the cause event and constitute a fair comparison. Most models perform worse on the 2-needle questions than the 1-needle questions. Fig. \ref{fig:1n_vs_2n} visualizes this pattern. 
This highlights the deficiency of using only 1-needle questions to evaluate long-video understanding.

\paragraph{Open-Source Models Exhibit Weaker World Modeling Ability.}

According to Tab.~\ref{tab:main_results}, open-source models generally perform worse than proprietary models. Specifically, proprietary models surpass all open-source models on both 1-needle and ID 2-needle questions. For VG 2-needle questions (Avg), all open-source models, except for Qwen2.5VL-7B and Aria-28B, perform at or below random levels, falling behind proprietary models. This gap may be due to insufficient human-behavior training data during the pretraining of open-source VLMs.


\paragraph{Performance Decreases with Increasing Needle Distance.} In Fig. \ref{fig:per_distance} (a) and (b), we present the VG 2-needle question performance (average order) for ChatGPT-4o and Gemini-1.5-pro. Performance is measured on questions with varying distances between the cause and effect events. The results show a clear decline as distance increases, especially when the evaluation aggregates answers from both parts. This phenomenon indicates that the distance between multiple needles has a significant impact when joint understanding of these needles is required. More details are provided in Appendix Sec.\ref{app:CE_length}. 

A possible confounder is the strength of causal relations\footnote{We thank an anonymous reviewer for suggesting this.}. That is, long distances between needles may be correlated with weak causal relationships. It may be the weak causality, rather than long distance, that causes performance decrease. To verify this conjecture, we test if humans can identify the cause clip over long distances. Two annotators answered 113 visual grounding 2-needle questions and the results with different needle distances are detailed in Fig. \ref{fig:per_distance} (c). 
We can observe that for 24 samples with a needle distance between 6 and 8 clips, the human performance is 87.5\% (Part 1), 87.5\% (Part 2) and 87.5\% (Both). This is even better than the performance on 97 samples with a distance between 3 and 5 clips: 83.5\% (Part 1), 86.6\% (Part 2) and 77.3\% (Both). These results demonstrate that humans can easily undestand the causal relations. Thus, we argue that the model performance degradation mainly stems from the long distances between needles.


\subsection{Pathological Behaviors of VLMs} \label{sec:2bias}
\paragraph{Positional Bias. } As shown in Tab. \ref{tab:main_results}, the accuracy of models responses varies depending on the positions of cause and effect clips. In VG 2-needle questions, Part 1 is grounded in the effect video clip, while Part 2 is grounded in the cause clip. In the forward-ordered clip sequence, the cause clip appears earlier, and all proprietary models show higher accuracy in Part 2 compared to Part 1; for instance, the accuracy of ChatGPT-4o exhibits an accuracy increase of 22.5\%. This result appears paradoxical, since from an information-processing perspective answering the Part 2 question requires an understanding of the video clip required for the Part 1 question. 
Why are models more accurate when locating the cause clip required by Part 2 than Part 1?

To answer this question, we reverse the order of the input video clips (the Reverse column of Tab. \ref{tab:main_results}). In this setting, Part 1 achieves much higher accuracy than Part 2.  This suggests that models pay more attention to clips appearing earlier; therefore, visual grounding for the cause clips is easier under normal playback, while reverse playback facilitates the visual grounding of the effect clips. To mitigate the influence of position bias, we evaluate model accuracy of VG 2-needle questions using the average of forward and reverse clip ordering.


\begin{figure}[t]
    \centering
    \includegraphics[width=0.8\textwidth]{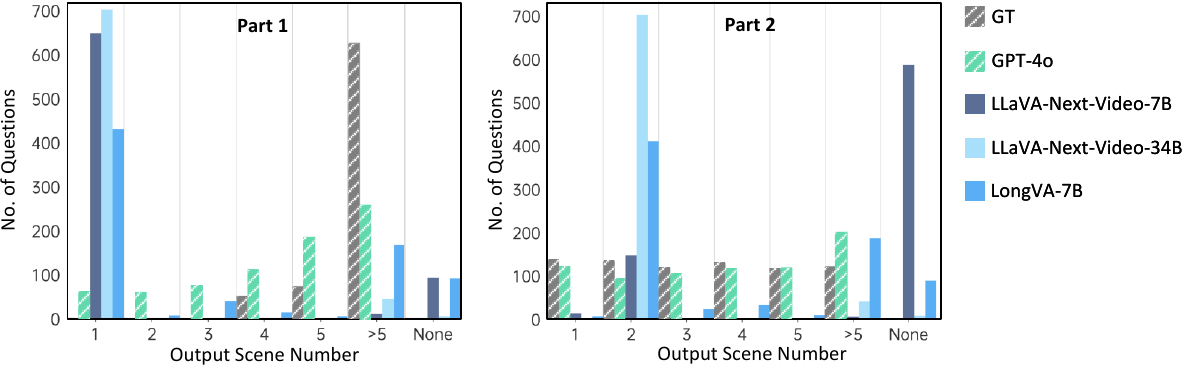}
    \caption{The answer distribution of various models in the forward evaluation of visual grounding 2-needle questions. GT denotes ground truth. None means no clip number is output. Predictions of opensource models are heavily concentrated in a few numbers, exhibiting significant bias.}
    \label{fig:distribution_bias}
\end{figure}

\paragraph{Static Output Bias. }
In Tab. \ref{tab:main_results}, we observe that some open-source VLMs achieve extremely low scores, sometimes zero, on VG 2-needle questions. When examined closely, we observe that some models tend to output the same prediction regardless of the question content. We call this phenomenon the static output bias. 

In Fig.~\ref{fig:distribution_bias}, we compare the output distributions of well-performing and poorly-performing models on VG 2-needle questions.
For example, the ground truth of Part 1 is mainly in clip 5 and beyond, as the effect clip appears after the cause. ChatGPT-4o shows a balanced distribution, aligning with its strong performance in Tab. \ref{tab:main_results}. 
In contrast, the three open-source models predominantly select clip 1 as the answer, demonstrating severe output bias. 
We experiment with different prompting strategies as attempts to fix this bias, but the models consistently produce fixed responses, indicating an inherent flaw in these models. The test prompts and output patterns are in Appendix Sec. \ref{app:prompts}.


\begin{table}[t]
\centering
\small
\caption{Results of models on different input settings. Numbers in parentheses indicate the absolute increase over random chance. 
1-N: 1-needle, 2-N: 2-needle, V: Video, N: Narration, Q: Question. When V is removed in causal 1-N, performance decreases to near chance level, demonstrating minimal textual bias. When V+N are removed in ID 2-N, performances stay above chance level but are still far from saturation.}
\vspace{0.2cm}
\label{tab:combined}
\resizebox{0.8\linewidth}{!}{
\begin{tabular}{lcccc}
\toprule
\multicolumn{1}{c}{\textbf{Models}} & \multicolumn{2}{c}{\textbf{Causal 1-N Questions}} & \multicolumn{2}{c}{\textbf{ID 2-N Questions}} \\

\cmidrule(lr){2-3} \cmidrule(lr){4-5}
& V+N+Q & N+Q & V+N+Q & Q only \\
\midrule
\multicolumn{5}{l}{\textit{Proprietary Models}} \\
Gemini-1.5-pro        & 35.6 (+25.7) & 12.3 (+2.4) & 60.9 (+35.9) & 39.0 (+14.0) \\
Claude-3-5-sonnet     & 26.5 (+16.7) & 10.7 (+0.9) & 60.5 (+35.5) & 29.2 (+4.2) \\
ChatGPT-4o            & 39.2 (+29.4) & 3.8 (-6.0) & 59.2 (+34.2) & 47.5 (+22.5) \\
\midrule
\multicolumn{5}{l}{\textit{Open-source Models}} \\
LongVA-7B             & 14.7 (+4.9) & 11.2 (+1.4) & 49.7 (+24.7) & 38.4 (+13.4) \\
InternVL2-8B          & 7.4 (-2.4) & 11.7 (+1.9) & 40.2 (+15.2) & 25.9 (+0.9) \\
LLaVA-Next-Video-7B   & 17.2 (+7.4) & 0.3 (-9.5) & 27.3 (+2.3) & 26.9 (+1.9) \\
\bottomrule
\end{tabular}
\vspace{-0.3cm}
}
\end{table}

\subsection{Analysis of Dataset Bias}
\paragraph{Textual Bias.} Textual bias refers to the extent that the dataset allows models to derive the correct answer from the accompanying text alone. 
To verify if our visual grounding test format effectively avoids textual bias, we test various models on causal 1-needle questions without providing the video clips. The results are presented in the first two columns of Tab. \ref{tab:combined}. After removing video input, model performance drops significantly to around random chance (9.8\%) or lower. For instance, the performance of ChatGPT-4o drops from 39.2\% to 3.8\%. These results indicate that our setting effectively mitigates textual bias, ensuring a reliable evaluation of multimodal understanding abilities. 


\paragraph{Knowledge Leakage from Pretraining.} To prevent performance underestimation caused by the out-of-distribution VG question format, we introduce an image description (ID) format. However, an VLM may have memorized details of movies from pretraining, which can be used to answer these questions. To assess the impact of prior knowledge, we test different models using question-only inputs. The results are shown in the last two columns of Tab. \ref{tab:combined}. We find that ChatGPT-4o, Gemini-1.5-pro, and LongVA achieve high performance on ID questions with question-only inputs, with +22.5\%, +14.0\%, and +13.4\% over random chance, respectively. This indicates that they can call upon prior knowledge to answer the questions. In contrast, Claude-3.5-sonnet and InternVL2-8B are less affected by prior knowledge. For almost all models tested, the performance with video and narration input is much higher than with question only, suggesting knowledge leakage alone is not sufficient for good performance. There remains substantial room for improvement (around 60–70\%) in \shortname, which could benefit from enhanced causal reasoning capability.

\newpage
\section*{Acknowledgments}
This research is supported by the RIE2025 Industry Alignment Fund – Industry Collaboration Projects (IAF-ICP) (Award I2301E0026), administered by A*STAR, by Alibaba Group and NTU Singapore through Alibaba-NTU Global e-Sustainability CorpLab (ANGEL), by the Nanyang Associate Professorship, and by the National Research Foundation Fellowship (NRFF13-2021-0006), Singapore.

%% file: sec/5.conclusion.tex


%% file: sec/9_appendix.tex
\section{Dataset Statistics}\label{app:stat}
\begin{figure}[t]
    \centering
    \includegraphics[width=0.8\textwidth]{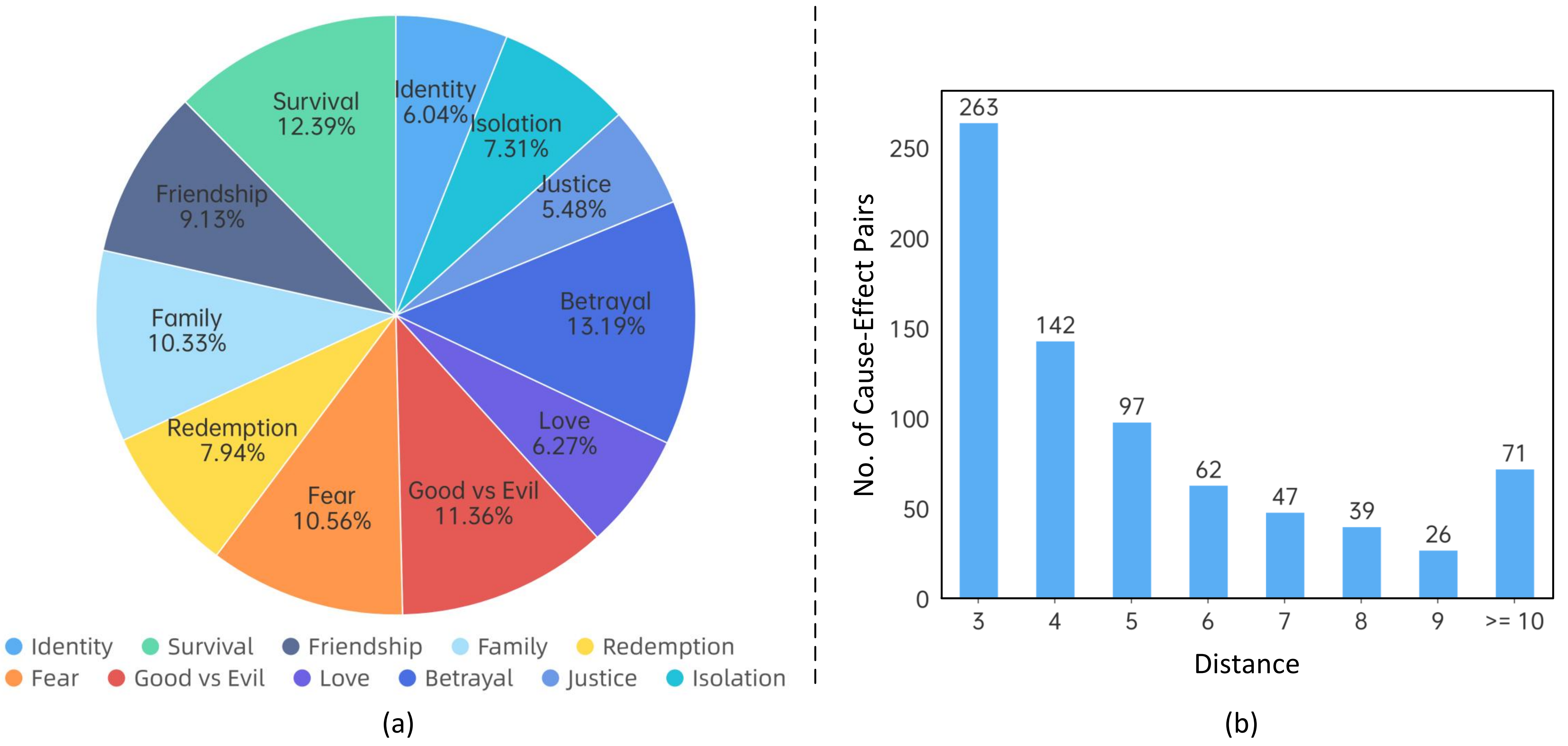}
    \caption{(a) Distribution of movie themes in \shortname{}. (b) Distribution of distance between cause and effect in \shortname{} }
    \label{fig:stats}
\end{figure}

The videos in \shortname{} are collected from the SyMoN \citep{symon} and YMS \citep{yms} datasets, comprising movie recap videos and their corresponding narrations (98 movies from SyMoN and 94 from YMS). After integration, the videos in \shortname{} have an average duration of 438 seconds, with 58.6 clips and 1,267.2 words per movie. The movie themes in our dataset are highly diverse. To quantify this diversity, we first use ChatGPT-4o to assign multi-labels to each narrative based on 11 predefined thematic categories, and then compute the frequency of each predicted label. The distribution is shown in Fig. \ref{fig:stats} (a). As detailed in Sec. \ref{sec:data_create}, we extract 747 cause-effect pairs from the narrations and construct 2-needle questions accordingly. We further analyze the distribution of their distances, defined as the number of clips between the cause and effect events, shown in Fig. \ref{fig:stats} (b).

\section{Event Graph Merging}\label{app: merge}
To obtain comprehensive and long-range causal relationships, we merge the extracted local and global event graphs. The merging process simply takes the union of the causal relations from different event graphs. For example, suppose the output of Graph A is [(1, 5), (3, 10), (21, 27)]. Here, (1, 5) denotes a causal relation from event 1 to event 5. Further suppose the output of Graph B is [(1, 5), (3, 10), (6, 11)]. The merged result is the union [(1, 5), (3, 10), (6, 11), (21, 27)].

\section{Evaluation Details}\label{app:GPUs}

\paragraph{Computing Resources.} We conduct experiments using four NVIDIA GeForce RTX A6000 GPUs for all open-source models. For proprietary models, we obtain results through their publicly available APIs. The inference batch size is set to 1.

\paragraph{Details of VLMs.} Current video understanding VLMs primarily rely on converting videos into sequences of images for processing. For example, LLaVA-Next-Video \citep{zhang2024video_llavanext} samples 32 frames from a video, encodes them into visual tokens using a Vision Transformer (ViT) \citep{dosovitskiy2020image_vit}, and feeds these tokens, along with textual instructions, into Vicuna-1.5 \citep{zheng2023judging_vicuna} to generate responses. Interleaved VLMs \citep{chen2024far_internvl, li2024aria} extend input formats to support text-image sequences.

Proprietary models such as Gemini-Pro \citep{team2024gemini}, GPT-4o \citep{openai2024chatgpt4o}, and Claude-3.5 \citep{anthropic2024claude3} benefit from substantially greater computational resources, enabling them to process millions of image tokens—achieving dense visual understanding at up to 2 frames per second. These models also support interleaved text-image inputs, offering stronger temporal and contextual reasoning capabilities over video content.

Next, we show the input content and format we used when evaluating different models.
\paragraph{Proprietary LLMs.} For GPT-4o, GPT-4o-mini, Gemini-1.5-Pro-002, and Claude-Sonnet, the input consists of:
(1) A series of movie scene images, including the scene containing the correct answer. Each scene is represented by five video frames stitched together vertically.
(2) The task instruction, which specifies the task type—either visual grounding or image description. In the visual grounding task, the model is instructed to identify the correct scene number, whereas in the image description task, the model is asked to select the correct option.
(3) The movie’s narration background.
(4) The question. If the task is image description, the input will also include textual answer options.
Thus, each input is structured as shown in the list below: 
\\
\textcolor{blue}{\tt<image>} \textcolor{blue}{\tt<image>} \textcolor{blue}{\tt<image>} ...\\
{\tt<text: Test Instruction>}\\
{\tt<text: Movie Narration>}\\
{\tt<text: Questions>}\\

\paragraph{Open-source VLMs. } For LLaVA-Next-Video-7B, LLaVA-Next-Video-34B, LLaVA-OneVision-7B, LongVU-7B, and LongVA-7B, the input formats are the same as those used for the proprietary models. However, these models utilize different chat templates during generation.
We follow the examples presented on each model's GitHub webpage or Hugging Face repository. \\
\noindent
LLaVA-Next-Video-7B:  \url{https://huggingface.co/llava-hf/LLaVA-NeXT-Video-7B-hf}\\
LLaVA-Next-Video-34B:  \url{https://huggingface.co/llava-hf/LLaVA-NeXT-Video-34B-hf}\\
LLaVA-OV-7B: \url{https://huggingface.co/llava-hf/llava-onevision-qwen2-7b-ov-hf}\\
LongVU-7B: \url{https://huggingface.co/Vision-CAIR/LongVU_Qwen2_7B} \\
LongVA-7B:  \url{https://github.com/EvolvingLMMs-Lab/LongVA}\\

The remaining three are VLMs pretrained on interleaved image-text data. To match their input format, we insert scene numbers between video segments. However, we do not insert any text between scenes, as this could provide shortcuts by overly aligning with the narration, rather than requiring true visual understanding.
\\
{\tt scene1}\textcolor{blue}{\tt<image>}{\tt scene2} \textcolor{blue}{\tt<image>}{\tt scene3}\textcolor{blue}{\tt<image>}\\
{\tt<text: Test Prompt>}\\
{\tt<text: Movie Narration>}\\
{\tt<text: Questions>}\\

The chat template is collected from: \\
InternVL2: \url{https://huggingface.co/OpenGVLab/InternVL2-8B}\\
Aria-28B: \url{https://huggingface.co/rhymes-ai/Aria}\\
Qwen2.5VL-7B: \url{https://huggingface.co/Qwen/Qwen2.5-VL-7B-Instruct}\\


\paragraph{Random Chance Calculation. }
For a 1-needle question, the model is required to select the correct scene from a sequence of given scenes (including extra padding scenes). Assuming the total number of input scenes is $N$, the random chance probability is $1/N$ (reported as a percentage).  For a visual grounding 2-needle question, where the model must make two separate selections, the random chance of answering both questions correctly is $1/N^{2}$.

Since the distance between the cause and effect scenes varies, the total number of input scenes $N$ differs for each question. Therefore, we compute the average input length $\bar{N}$ over each type of questions and determine the random chance probability. For 1-needle questions, the random chance probability is 9.8\%.  For visual grounding 2-needle questions, the probability of correctly identifying a single part is 9.8\%, while the probability of getting both correct is only 1.0\%.  For image description 2-needle questions with four options, the random chance is 25.0\%.

\section{Additional Experiments}\label{app:CE_length}
In this section, we provide a deeper analysis of several aspects: (1) the influence of the distance between cause and effect events; (2) textual grounding versus visual grounding; (3) the uniqueness of answers to Part 2 questions; (4) existence of counterfactual answers in ID 2-needle questions; and (5) non-stacked versus stacked image input settings.

\paragraph{The Influence of Needle Distance.} 
\begin{figure}[t]
    \centering
    \includegraphics[width=0.7\textwidth]{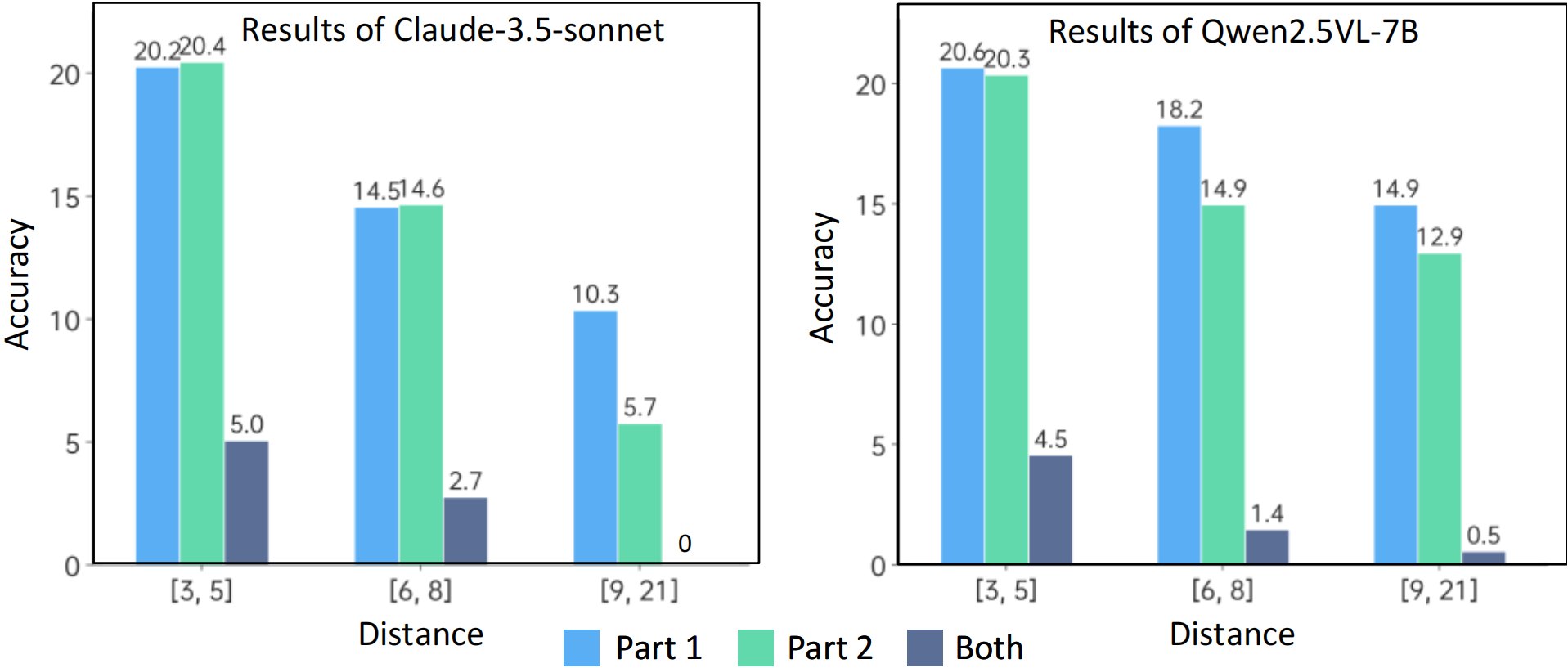}
    \caption{The 2-needle visual grounding performance of various models on questions with different needle distances.}
    \label{fig:per_dis_for_rev}
\end{figure}

Fig.~\ref{fig:per_distance}~(b) in Sec.~\ref{sec:main_results} shows that the 2-needle visual grounding performance decreases as the distance between the cause and effect events increases, based on evaluations with two proprietary models. Here, we provide additional results in Fig. \ref{fig:per_dis_for_rev}. These results indicate that needle distance significantly impacts model performance in multi-needle questions.

\paragraph{Textual vs. Visual Grounding.}

\begin{figure}[t]
    \centering
    \includegraphics[width=0.8\textwidth]{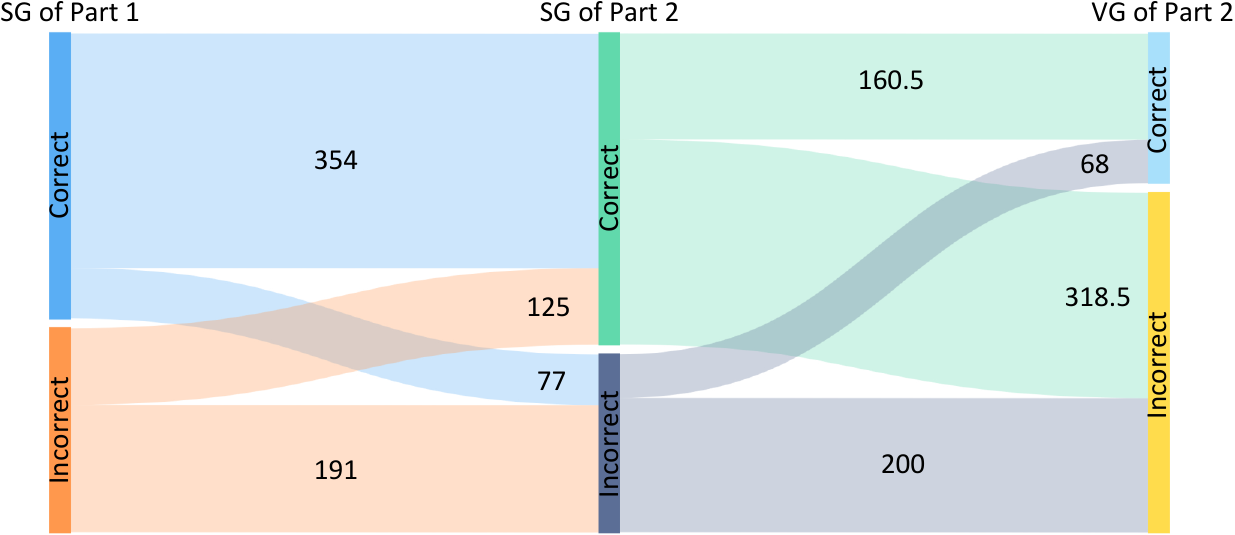}
    \caption{Performance of ChatGPT-4o on the sentence grounding (SG) task and the visual grounding (VG) task on the 2-needle questions. In the SG tasks, the sentences are always arranged in their natural order. In the VG task, we take the average of the forward and reverse orderings.}
    \label{fig:erro_analysis}
\end{figure}

To better understand the mechanisms underlying the models’ performance on VG 2-needle questions, we conduct additional experiments to analyze visual and textual grounding. We introduce a new task called Sentence Grounding (SG), which requires the model to select the sentence that contains the answer from the story narration text, which are part of the input of \shortname. We compare the SG performance of GPT-4o on Part 1 and Part 2 questions, and summarize the results in Fig. \ref{fig:erro_analysis}. We make the following observations. 

First, the results show that the Part 2 question relies on joint understanding of both the cause and the effect sentences. We compute the conditional probability that GPT-4o answers SG Part 2 correctly given that it answers SG Part 1 correctly, $\mathrm{P}(\text{SG-Part2} = {\text{correct}} \mid \text{SG-Part1} = {\text{correct}}) = 82.1\%$. In contrast, the conditional probability that GPT-4o answers SG Part 2 correctly given that it answers SG Part 1 incorrectly, $\mathrm{P}(\text{SG-Part2} ={\text{correct}} \mid \text{SG-Part1} = {\text{incorrect}}) = 39.6\%$. The large gap between the two conditional probabilities demonstrates that performance on SG Part 2 significantly depends on SG Part 2. This suggests the existence of a sequential dependency between the two parts. The finding demonstrates that, at least in the textual modality, the two-needle questions benefit from the joint understanding of both needles.

Second, the dependence of VG performance on SG performance is limited. We compute the conditional probability that GPT-4o answers VG Part 2 correctly given that it answers SG Part 2 correctly, $\mathrm{P}(\text{VG-Part2 = \text{correct}} \mid \text{SG-Part2} = {\text{correct}}) = 33.5\%$. In contrast, the conditional probability that the model answers VG Part 2 correctly given that it answers SG Part 2 incorrectly is $\mathrm{P}(\text{VG-Part2} = {\text{correct}} \mid \text{SG-Part2} = {\text{incorrect}}) = 25.4\%$. The difference between the two probabilities is small. This shows the correctness of SG Part 2 only makes minor contribution to VG Part 2. That is, even if the model is wrong on SG Part 2, it still has a decent chance of being correct on VG Part 2, relative to the alternative situation that the model is correct on SG Part 2. Hypothetically, a main reason for this phenomenon is that the model utilizes a separate mechanism for visual reasoning, which operates more or less independently of reasoning over the textual modality. 

Lastly, on SG Part 2, GPT-4o achieves 63.7\% accuracy, much higher than the 30.3\% on VG Part 2, suggesting the model is better at textual than visual reasoning. This large gap indicates substantial room for improvement in visual comprehension of VLMs.

\paragraph{Answer Uniqueness of  Part 2 Questions.} 
We conduct an experiment to verify if Part 2 questions have unique answers. In causal relationship extraction, an event graph may contain multiple nodes pointing to the same node, indicating that a single event can have multiple causes. This may introduce multiple ground truth answers to Part 2 questions, as they require retrieving the cause of an event. 


Our experiment is designed as follows: among the $747$ causal relationships of 2-needle questions, we identify $99$ instances where the effect sentence is not uniquely paired with a single cause sentence. For each effect sentence in them, we feed the corresponding Part 2 question and cause sentences (on average, $2.02$ candidate causes per effect) to ChatGPT-4.1. Then, it is asked to select the cause sentence that contains the answer to the Part 2 question. The results show that out of the $99$ such cases, ChatGPT-4.1 selects the wrong cause sentence in only one instance. This indicates that the answer to Part 2 question is mostly unique. 

\begin{table}[t]
\centering
\small
\caption{Comparison of results between stacked image inputs and non-stacked image inputs. }
\label{tab:satcked}
\resizebox{\linewidth}{!}{
\begin{tabular}{@{}ccccccccccc@{}}
\toprule
\multirow{3}{*}{\textbf{Models}}         & \multirow{3}{*}{\begin{tabular}[c]{@{}c@{}}\textbf{Input}\\ \textbf{Setting}\end{tabular}} & \multicolumn{9}{c}{\textbf{VG 2-Needle Questions}}                                           \\ \cmidrule(l){3-11} 
                                &                                                                          & \multicolumn{3}{c}{Forward} & \multicolumn{3}{c}{Reverse} & \multicolumn{3}{c}{Avg} \\ \cmidrule(l){3-11} 
                                &                                                                          & Part 1   & Part 2   & Both  & Part 1   & Part 2   & Both  & Part 1  & Part 2 & Both \\ \midrule
\multirow{2}{*}{ChatGPT-4o}     & non-stacked                                                              & 5.0      & 29.4     & 1.4   & 29.7     & 4.1      & 0.9   & 17.4    & 16.8   & 1.2  \\
                                & stacked                                                                  & \textbf{16.7}     & \textbf{39.2}     & \textbf{9.4}   & \textbf{45.4}     & \textbf{21.2}     & \textbf{13.4}  & \textbf{31.1}    & \textbf{30.2}   & \textbf{11.4} \\ \midrule
\multirow{2}{*}{Gemini-1.5-pro} & non-stacked                                                              & 5.4      & 37.3     & 2.9   & 26.3     & 17.5      & 6.1   & 15.8    & 27.4   & 4.5  \\
                                & stacked                                                                  & \textbf{21.0}     & \textbf{40.0}     & \textbf{10.2}  & \textbf{35.7}     & \textbf{21.4}     & \textbf{8.4}   & \textbf{28.4}    & \textbf{30.7}   & \textbf{9.3}  \\ \midrule
\multirow{2}{*}{Qwen2.5VL-7B} & non-stacked                                                              & 2.1      & 11.0     & 0.0   & \textbf{13.8}     & 4.4      & 0.4   & 8.0    & 7.7   & 0.2  \\
                                & stacked                                                                  & \textbf{27.6}     & \textbf{17.7}     & \textbf{5.0}  & 11.2     & \textbf{18.9}     & \textbf{1.9}   & \textbf{19.4}    & \textbf{18.3}   & \textbf{3.4}
\\
\bottomrule
\end{tabular}
\vspace{-0.3cm}
}
\end{table}


\paragraph{Existence of Counterfactual Answers in ID 2-Needle Questions.} A counterfactual answer refers to an answer that is supported apparently by the input video and can be ruled out only by causal reasoning. If the video contains only the correct answer and no counterfactual answers, a neural network can easily rule out wrong answers without locating the required video clip. 

For example, in Fig. \ref{fig:framework}, the image description question is ``What tragedy was commemorated by the memorial service in Metropolis? When the tragedy that causes the memorial service occurs, what color were the energy blasts emanating from the creature? A. Yellow B. Blue C. Red D. White.'' The correct answer is ``Yellow'', which appears in the cause clip. ``Blue'' is a competing answer as a blue light is shown in another battle clip at 2'37 of the video. Due to the presence of both answers in the input video, to correctly answer the question, the model must identify the cause clip through causal reasoning.

We adopt two question design methods to ensure the existence of counterfactual answers in image description questions. First, we are deliberately ambiguous about the reference in the question wording, as discussed in Sec. \ref{method:2n-gen}. Second, we encourage the LLM to search for visually similar but potentially confusing content from adjacent clips, using the instructions in Appendix Fig. \ref{fig:vis-2-needle}. 

To demonstrate the existence of counterfactual answers, we conduct an experiment with the following setting: (1) We remove the cause clip from the input and any information that helps to locate the cause clip from the question, allowing the model to answer the question using other clips. For example, the aforementioned question becomes `` What color were the energy blasts emanating from the creature?''; (2) We replace the ground truth answer with ``None is correct,'' so if there are no counterfactual answers, the model should choose this option. In experiments with ChatGPT-4o, the model selects ``None is correct'' for only 17.8\% of questions. This suggests that most image description questions in \shortname{} have counterfactual answers.


\paragraph{Non-stacked vs. Stacked Image Input. }
We explain the rationale for the design choice of inputting each video clip by stacking the video frames into a single image.  We represent a video clip using five uniformly sampled frames. After that, we stack these frames from top to bottom as a single image as input to the VLMs. There are two reasons for the stacked approach. First, most open-source models have a limited window size and can only process 32 frames at a time. Therefore, without stacking, processing multiple video clips would quickly exceed the window size. Second, for proprietary models that can accept more frames as input, stacking the frames can help the model better capture and comprehend the story within a video clip. As verification, we conduct additional experiments on VG 2-needle questions using ChatGPT-4o, Gemini-1.5-pro and Qwen2.5VL-7B with non-stacked frames as input. For a fair comparison, we instruct the model to output the frame number containing the answer and calculate the clip number by dividing the frame number by five, as every five frames belong to a single clip. The results are detailed in Tab. \ref{tab:satcked}. We can observe that stacked frames improve performance significantly, indicating that the stacked method enhances visual understanding.

\section{Limitations}\label{app:limits}
While our benchmark addresses several limitations of prior work, such as the lack of multi-needle evaluation on long-context video understanding and the understanding of human behaviors, there remain areas for future improvement. 

First, our videos are primarily sourced from movie clips, which provide a rich set of event causal relations. Compared to real life, movies tend to over-emphasize uncommon events, such as serendipity,  intricate conspiracy, or dramatic conflicts. These events could serve as effective out-of-distribution tests for models primarily trained on common events such as ego-centric videos. However, they should not be taken as necessarily representative of real-life events. Second, our current evaluation focuses on VLMs. Many of them, particularly open-source ones, do not yet support audio inputs. Accordingly, we exclude audio from our inputs, but future extensions of the benchmark will incorporate audio to enable evaluation of more comprehensive multimodal understanding.

\section{Generation and Evaluation Prompts}\label{app:prompts}
We present prompting templates for causal-relationship extraction (Fig. \ref{fig: causal_extraction}),  1-needle question generation (Fig. \ref{fig:cau-1-needle}), visual-grounding 2-needle question generation (Fig. \ref{fig:cau-2-needle}), and image-description 2-needle question generation (Fig. \ref{fig:vis-2-needle}). Visual-grounding 2-needle questions and 1-needle questions are generated with GPT-o1 (openai-o1-preview-2024-09-12), with sentence rephrasing performed by Gemini-1.5-Pro-002 as part of the visual-grounding 2-needle pipeline; image-description 2-needle questions are generated with Gemini-1.5-Pro-002.

In Fig. \ref{fig:vis-2-needle-options}, we continue using the same model to generate answer choices. At this stage, to increase the difficulty of the options, we input the cause clip as well as the two adjacent frames and prompt the model to consider generating more challenging options based on objects appearing in the remaining scenes.

\begin{tcolorbox}[
  colback=white,
  colframe=skyblue!80!black,
  title=Causal Relationships Extraction (GPT-o1-preview-2024-09-12),
  coltitle=white,
  fonttitle=\bfseries,
  colbacktitle=skyblue!90!black,
  boxrule=0.6pt,
  width=0.95\textwidth
]
  
Here is a list of nodes (events) from a story event graph. We want you to fill
in the edges of the event graph with causal connections between nodes. An event
graph contains nodes and edges. Each node represents an event, and each edge
represents the causal connection between two events.

\medskip
\textbf{Example Input:} \\ 
Node 0: When Dan goes to school in the morning, he has to take the bus.\\
Node 1: One day Dan was running late, and missed the bus to school.\\
Node 2: Dan called his friend Pete, and asked for a ride to school.\\
Node 3: Pete gave Dan a ride to school, but Dan was late for his first class.\\
Node 4: Luckily Dan wasn’t late for any of his other classes that day.

\medskip
\textbf{Example Output:} \\ 
Edge 0: (Node 0 -> Node 1)\\
Edge 1: (Node 1 -> Node 2)\\
Edge 2: (Node 2 -> Node 3)\\
Edge 3: (Node 1 -> Node 3)\\
Edge 4: (Node 3 -> Node 4)\\
(continue with another five demonstrations)

\medskip
Now, it is your turn to construct the event graph for the following event list.\\
Event List:\\
Node 0: \texttt{<S1>}\\
Node 1: \texttt{<S2>}\\
Node 2: \texttt{<S3>}\\
Node 3: \texttt{<S4>}\\
Node 4: \texttt{<S5>}\\
\textbf{Output:}

\end{tcolorbox}
\vspace{-0.5em}
\captionof{figure}{The prompt for causal relationships extraction.}
    \label{fig: causal_extraction}

\begin{tcolorbox}[
  colback=white,
  colframe=skyblue!80!black,
  title=1-Needle Question Generation (GPT-o1-preview-2024-09-12),
  coltitle=white,
  fonttitle=\bfseries,
  colbacktitle=skyblue!90!black,
  boxrule=0.6pt,
  width=0.95\textwidth
]
\textbf{Caulsal 1-Needle Question}

\medskip
Given a pair of sentences with a causal relationship, create a question about the ``Effect" sentence such that the answer to this question 
is found exclusively in the ``Cause" sentence. Ensure that the answer does not appear within the “Effect” sentence itself.

\medskip
\textbf{Example Input:}  
Cause: Blaming himself, Harry Hart, codenamed Galahad, delivers a medal for valor to Lee’s widow, Michelle and his young son, Eggsy, 
saying that if they ever need help, they should call the phone number on the back of the medal.  
Effect: Arrested for stealing a car, Eggsy calls the number on the medal.

\medskip
\textbf{Example Output:}  
Why did Eggsy have a phone number to call when he was arrested for stealing a car?  
(continue with another four demonstrations)

\medskip
\textbf{Input:}  \\
Cause: \texttt{<Cause Sentence>} \\
Effect: \texttt{<Effect Sentence>}

\medskip
\textbf{Output:}\\
\medskip
\noindent\makebox[\linewidth]{\dotfill}
\medskip
\textbf{Noncausal 1-Needle Question}\\
Given a sentence, please use the content of it as context to create a question whose answer appears in the sentence.
Please output the question and answer in the following format, keeping the answer as concise as possible:
\\Question: \texttt{<Question>}\\
Answer: \texttt{<Answer>}\\

\medskip
\textbf{Input:}  \\
Sentence: \texttt{<Cause Sentence/Effect Sentence>} \\

\medskip
\textbf{Output:}

\end{tcolorbox}

\captionof{figure}{The prompt for generating 1-needle questions.}
    \label{fig:cau-1-needle}
    
\begin{tcolorbox}[
  colback=white,
  colframe=skyblue!80!black,
  title=Visual Grounding 2-Needle Question Generation (Step 1)\\ (Gemini-1.5-pro-002),
  coltitle=white,
  fonttitle=\bfseries,
  colbacktitle=skyblue!90!black,
  boxrule=0.6pt,
  width=0.95\textwidth
]

\textbf{Step 1: Rephrase the cause and effect sentences}

You have a pair of sentences: one indicating a cause and the other describing its effect. 
Each sentence has a related image that complements the content expressed in the sentence.
Your goal is to rephrase both sentences so that the causal relationship between them is explicitly clear.
Please ensure that after rephrasing, there is a piece of shared information present in both sentences that establishes the causal relationship between them.
Remember that the rephrased sentences should not include any content that is not present in the original text or images.

\medskip
\textbf{Example 1:}\\
\textbf{Input:} \\
Cause: Gordon, who was actually sent to save Rachel, is unable to make it there in time and Rachel dies. \\
\texttt{<insert cause scene here>}\\
Effect: The Joker is able to locate Dent in a hospital and manipulates him into seeking revenge for Rachel’s death.\\
\texttt{<insert effect scene here>}\\
\smallskip
\textbf{Output:} \\
Cause: Gordon, tasked with rescuing Rachel, arrives too late, and she is dead. \\
Effect: The Joker confronts Dent in a hospital and manipulates him into seeking revenge for Rachel’s death against Gordon.\\

\textbf{Example 2:}\\
\textbf{Input:} \\
Cause: The remaining robber reveals himself to be the joker, a crazed supervillain, and escapes the bank in a school bus. \\
\texttt{<insert cause scene here>}\\
Effect: Several mob leaders hold a conference, which is interrupted by the joker.\\
\texttt{<insert effect scene here>}\\
\smallskip
\textbf{Output:} \\
Cause: After the final robber reveals himself to be the joker---an unstable supervillain---and escapes the bank by blending into a line of school buses, the criminal underworld is thrown into disorder. \\
Effect: To restore order, several prominent mob leaders convene a clandestine meeting, only to have the joker himself crash their conference.

\smallskip
\textbf{Input:}\\
Cause: \texttt{<Cause Sentence>} \texttt{<Cause Scene Images>}\\
Effect: \texttt{<Effect Sentence>} \texttt{<Effect Scene Images>}\\
\end{tcolorbox}
\captionof{figure}{The prompt for generating visual grounding 2-needle questions.}

\begin{tcolorbox}[
  colback=white,
  colframe=skyblue!80!black,
  title=Visual Grounding 2-Needle Question Generation (Step 2)\\ (GPT-o1-preview-2024-09-12),
  coltitle=white,
  fonttitle=\bfseries,
  colbacktitle=skyblue!90!black,
  boxrule=0.6pt,
  width=0.95\textwidth
]
\textbf{Step 2: Generate questions based on bridge entities}

You have a pair of sentences: one indicating a cause and the other describing its effect. Your goal is to create a sentence containing two questions. Please complete this goal step by step, as shown in the examples. (Please follow the output format shown in the example.)

\medskip
\textbf{Example Input:}\\
Cause: Feeling responsible, Harry Hart, known as Galahad, gives a medal for valor to Lee’s widow, Michelle, and his young son, Eggsy, instructing them to call the number on the medal if they ever need help.
Effect: When Eggsy is arrested for stealing a car, he calls the number on the medal for help.

\textbf{Example Output:}\\
\textbf{Step 1:} \\
Instruction: Identify a piece of shared information that appears in both “Cause” and “Effect” sentences and establishes their causal relationship. \\
Output: The shared information is `call the number on the medal'.\\
\textbf{Step 2:} \\
Instruction: Create a question that uses the “Effect” sentence as context and shared information as the answer. Rephrase the shared information to be less specific. \\
Output: First, according to the context, `call the number on the medal' can be rephrased into `an approach'. Then, the question is `What approach did Eggsy take to seek help when he was arrested for stealing a car?'\\
\textbf{Step 3:} \\
Instruction: Based on the previous question, create another question using the “Cause” sentence. The answer should be the shared information.\\
Output: The question is `Why was he able to adopt such an approach?'\\
\textbf{Step 4:} \\
Instruction: Combine the two questions into one sentence.\\
Output: What approach did Eggsy take to seek help when he was arrested for stealing a car, and why was he able to adopt such an approach?\\
\medskip
\textbf{Input:}\\
Cause:  \texttt{<Rephrased Cause Sentence>}\\
Effect:  \texttt{<Rephrased Effect Sentence>}\\
Output:
\end{tcolorbox}
\vspace{-1em}
\captionof{figure}{The prompt for generating visual grounding 2-needle questions (continued).}
\label{fig:cau-2-needle}


\begin{tcolorbox}[
  colback=white,
  colframe=skyblue!80!black,
  title=Image Description 2-Needle Question Generation (Step 1-2)\\ (Gemini-1.5-pro-002),
  coltitle=white,
  fonttitle=\bfseries,
  colbacktitle=skyblue!90!black,
  boxrule=0.6pt,
  width=0.95\textwidth
]

\textbf{Step 1:}  
You are given a question and its corresponding context. Your task is to concisely identify the following:

The question asks for specific information about an event. Identify this event (which should match the effect event mentioned in 
the context) and summarize it in a short phrase as \textbf{Effect Event}.  
Then, in the cause sentence, identify the cause event that leads to the effect event found in the first step, and summarize it in a 
short phrase as \textbf{Cause Event}.

\medskip
\textbf{Output:} \\
Cause Event: \texttt{<cause event>} \\
Effect Event: \texttt{<effect event>}

\medskip
\textbf{Input:} \\
Context: \texttt{<context>} \\
Question: \texttt{<part1>}

\tcblower
\textbf{Step 2:}  
You are given a context that can answer a two-part question concerning the same concept (the ``bridge entity")—though it may be 
referred to as a ``mission," ``event," ``approach," or another term in the question. Identify the bridge entity in the two-part question and 
clarify it based on the context. It often follows ``what" or ``which".\\
\smallskip
\textbf{Example:}  \\
Question: What event caused Bryant to order Deckard to retire the replicants, and what is Deckard's profession that relates to such an 
event?  \\
Context: Deckard, a Blade Runner tasked with "retiring" replicants, is informed that four—Leon, Roy Batty, Zhora, and Pris—have 
illegally arrived on Earth. Due to the arrival of the four replicants (Leon, Roy Batty, Zhora, and Pris), Bryant orders Deckard to retire them.  \\

Output: \\
Bridge Entity: the event \\
Reference: arrival of the four replicants
\end{tcolorbox}
\captionof{figure}{The prompt for generating image description 2-needle questions.}

\begin{tcolorbox}[
  colback=white,
  colframe=skyblue!80!black,
  title=Image Description 2-Needle Question Generation (Step 3-4)\\ (Gemini-1.5-pro-002),
  coltitle=white,
  fonttitle=\bfseries,
  colbacktitle=skyblue!90!black,
  boxrule=0.6pt,
  width=0.95\textwidth
]
\textbf{Step 3:}  
You are given five images. Your task is to generate a specific visual question with an answer that fits the following format:

``When the event that causes \texttt{<effect event>} occurs, \texttt{<visual question>}? Answer: \texttt{<a short answer to the visual question>}."

\textbf{Guidelines:}\\  
- Ensure the visual question is specific—it should focus on aspects such as environment, object attributes, facial expressions, or clothing.\\  
- Avoid vague questions like ``What is the significant object?"\\  
- Avoid well-known facts such as ``What is the color of Superman’s suit?" \\ 
- Avoid overly obvious questions like ``What is the doctor wearing?" if the answer is trivial.\\  
- Instead, ask detailed and challenging questions like:\\  
  ``What color is the jacket he is wearing?"\\  
  ``How is the character’s facial expression?"  \\
  ``What objects are placed on the table?" \\ 
  ``What is the lighting condition in the background?"\\

\medskip
\textbf{Output:}  
When the event that causes \texttt{<EFFECT EVENT>} occurs, ...

\tcblower
\textbf{Step 4:}  
You are given a question that asks about Event B, which leads to Event A, meaning it contains references to two events.

Your task is to rewrite the question so that it reads naturally while ensuring the following modifications:  
- Replace any mention of ``\texttt{<REFERENCE>}" in the question with its corresponding bridge entity: ``\texttt{<BRIDGEENTITY>}".  
- Ensure the rewritten question is grammatically correct and sounds natural.  
- Maintain the focus on Event B (do not rewrite the question to ask only about Event A).

\medskip
\textbf{Input:}  Question: \texttt{<question>}

\medskip
\textbf{Output:}  Rewrite question:

\end{tcolorbox}
\vspace{-0.5em}
\captionof{figure}{ The prompt for generating image description 2-needle questions (continued).}
\label{fig:vis-2-needle}

\begin{tcolorbox}[
  colback=white,
  colframe=skyblue!80!black,
  title=Image Description 2-Needle Question Options Generation\\ (Gemini-1.5-pro-002),
  coltitle=white,
  fonttitle=\bfseries,
  colbacktitle=skyblue!90!black,
  boxrule=0.6pt,
  width=0.95\textwidth
]
  
You will be given three images along with a question that has a correct answer.  
Your task is to generate three additional challenging answer choices.

\medskip
\textbf{How to Generate Challenging Options:} \\ 
- If the correct answer involves attributes like color, pattern, shape, or emotion, create alternative choices with different values within the same category.\\
- Increase difficulty by using existing objects that appear in the images.\\

\medskip
\textbf{Output format:}  
\texttt{<Question>}\\
A. \texttt{<Option A>}  B. \texttt{<Option B>}  C. \texttt{<Option C>}  D. \texttt{<Option D>}\\
Correct Answer: \texttt{<Letter>}

\medskip
\textbf{Input:}  
Question and its correct answer: \texttt{<Question>}

\medskip
\textbf{Output:}

\end{tcolorbox}
\vspace{-0.5em}
\captionof{figure}{The prompt for generating options for image description 2-needle questions.}
    \label{fig:vis-2-needle-options}



\paragraph*{Question Quality Evaluation Prompts}\label{app: question_eval_prompts}
We present the prompts used for evaluating the question quality (Bridge Entity, Semantic Coherence, and Readability) in Fig. \ref{fig:quesiton_quality_prompt}

\begin{tcolorbox}[
  colback=white,
  colframe=skyblue!80!black,
  title=Question Quality Validation Prompts,
  coltitle=white,
  fonttitle=\bfseries,
  colbacktitle=skyblue!90!black,
  boxrule=0.6pt,
  width=0.95\textwidth
]

\textbf{Factor 1} \\
You are given two causally related sentences and a bridge entity. \\
You need to finish the following task:  
determine whether this bridge entity is a piece of shared information that establishes the causal relationship between the two sentences. You should answer `yes' or `no'.  

\medskip
Cause: \texttt{<Cause>} \\
Effect: \texttt{<Effect>} \\
Bridge Entity: \texttt{<Bridge Entity>}

\noindent\makebox[\linewidth]{\dotfill}
\medskip
\textbf{Factor 2} \\
You are given a bridge entity and its vague reference. \\
You need to finish the following task:  
determine whether this vague reference is more ambiguous than the original bridge entity but still preserves its core meaning. You should answer `yes' or `no'.  

\medskip
Bridge Entity: \texttt{<Bridge Entity>} \\
Vague Reference: \texttt{<Vague Reference>}

\noindent\makebox[\linewidth]{\dotfill}
\medskip
\textbf{Factor 3} \\
You are given two causally related sentences and a question. \\
Please evaluate the \emph{semantic coherence} of the question w.r.t the sentences —  
that is, whether the question matches the facts and events described in both sentences  
and does not introduce unrelated or contradictory content.  

\medskip
You can only output a score from 1 to 5, where 5 = excellent semantic coherence.

\medskip
Cause: \texttt{<Cause>} \\
Effect: \texttt{<Effect>} \\
Question: \texttt{<Question>}

\noindent\makebox[\linewidth]{\dotfill}
\medskip
\textbf{Factor 4} \\
You will be given a question.  
Please rate its readability on a scale from 1 to 5, where 5 means very readable and 1 means very not readable.  

Consider the following criteria: Naturalness; Grammar; Clarity.  
You can only output one number.

\medskip
Question: \texttt{<Question>} \\
Score:

\end{tcolorbox}
\vspace{-0.5em}
\captionof{figure}{The prompt used for evaluating the quality of generated questions.}
    \label{fig:quesiton_quality_prompt}


\paragraph*{Cause-Effect Uniqueness Test Prompts}
We present the prompts used to test the uniqueness of the generated questions with respect to multiple candidate cause sentences in Fig. \ref{fig:unique_prompt}

\paragraph*{Model Performance Evaluation Prompts}\label{app: eval_prompts}

We present the task instruction prompts used for testing the models in Fig. \ref{fig:test_prompt}

For visual grounding 2-needle questions, there are two modes: Forward playback and Reverse playback. 
This distinction is explicitly stated in the instruction prompt.

\paragraph*{Various Instruction Templates and Model Output Examples}\label{app:append_outputbias}
In the Sec.~\ref{sec:2bias} of the main text, we observed that many open-source models exhibit severe output bias. Specifically, when testing the visual grounding 2-needle questions, these models tend to produce the same answer (``Scene 1 for part 1 and Scene 2 for part 2.") across different questions. To investigate whether this bias was caused by the prompt, we tested the models with different variations of the prompt. However, we still found that they continued to generate highly consistent responses. Below, in Fig.\ref{fig:different_prompt_2}, we list the most common outputs observed. Each type of response accounts for a significant proportion of the models’ answers across different questions. We found that although the model's outputs varied slightly with different prompts, a significant portion of the responses remained completely identical.

\begin{tcolorbox}[
  colback=white,
  colframe=skyblue!80!black,
  title=Prompt for Uniqueness Test and Sentence Grounding,
  coltitle=white,
  fonttitle=\bfseries,
  colbacktitle=skyblue!90!black,
  boxrule=0.6pt,
  width=0.95\textwidth,
]

\textbf{Uniqueness Test} \\
You are given two questions, the second question is based on the first question.  
You are also provided with the sentence containing the answer to the first question.  
Please identify which sentence contains the answer to the second question.

\medskip
Question 1: \texttt{<Question 1>} \\
Answer to Question 1: \texttt{<Answer to Q1>} \\
Question 2: \texttt{<Question 2>} \\

Answer Candidates to Question 2: \\ 
0: \texttt{<Candidate 0>} \\
1: \texttt{<Candidate 1>} \\

Index Number of the Answer: 

\medskip
\noindent\makebox[\linewidth]{\dotfill}
\medskip

\textbf{Sentence Grounding} \\
You are given a movie context and a question about it.  
Identify two sentences to answer the following question, which consists of two parts.

\medskip
Note: First, indicate which sentence contains the answer to Part 1, and then specify which sentence contains the answer to Part 2.  
Your answer format should be: `Sentence \texttt{<number>} for Part 1 and Sentence \texttt{<number>} for Part 2.'

\medskip
Context: \texttt{<Context>} \\
Question: \texttt{<Question>}

\end{tcolorbox}
\vspace{-0.5em}
\captionof{figure}{The prompt for testing the uniqueness.}
    \label{fig:unique_prompt}

\begin{tcolorbox}[
  colback=white,
  colframe=skyblue!80!black,
  title=Task Instruction Prompts,
  coltitle=white,
  fonttitle=\bfseries,
  colbacktitle=skyblue!90!black,
  boxrule=0.6pt,
  width=0.95\textwidth
]
\textbf{1-Needle Questions} \\ 
You are given a movie context consisting of several sentences and a series of consecutive scenes from the movie.  
Each scene is composed of five images stitched together from top to bottom.  
Identify which scene contains the necessary clues to answer the following question.  

Note: Please provide the index number of the scene (e.g., 1, 2, or 3) that contains the necessary information.

Context: \texttt{<Context>} \\
Question: \texttt{<Question>}

\noindent\makebox[\linewidth]{\dotfill}
\medskip
\textbf{Visual Grounding 2-Needle Questions (Forward)} \\ 
You are given a movie context and a sequence of consecutive movie scenes,  
each composed of 5 images stacked vertically. Identify two scenes to answer the following question, which consists of two parts.  \\
\textbf{Visual Grounding 2-Needle Questions (Reverse)}\\ 
You are given a movie context and a sequence of reverse-order consecutive movie scenes,  
each composed of 5 images stacked vertically. Identify two scenes to answer the following question, which consists of two parts.  

Note: First, indicate which scene contains the answer to Part 1, and then specify which scene contains the answer to Part 2.  
Your answer format should be: `Scene \texttt{<number>} for Part 1 and Scene \texttt{<number>} for Part 2.'

Context: \texttt{<Context>} \\
The question has two parts: \\
Part 1: \texttt{<Part1>} \\
Part 2: \texttt{<Part2>}

\noindent\makebox[\linewidth]{\dotfill}
\medskip
\textbf{Image Description 2-Needle Questions} \\ 
You are given a movie context consisting of several sentences and a series of consecutive scenes from the movie.  
Each scene is composed of five images stitched together from top to bottom.  

You will be asked a two-part question.  
The first part of the question is designed to help you identify the scene that contains the necessary information to answer the second part.  
Using the correct scene, accurately answer Part 2 by selecting the most precise answer from the four given options.  

Note: Please answer only the second part of the multiple-choice question.

Context: \texttt{<Context>} \\
Question: \texttt{<Question>}

\end{tcolorbox}
\leavevmode\vspace{-0.5em}\captionof{figure}{The evaluation prompts for different types of questions.}
    \label{fig:test_prompt}

\begin{tcolorbox}[
  colback=white,
  colframe=skyblue!80!black,
  title=Prompts and Outputs,
  coltitle=white,
  fonttitle=\bfseries,
  colbacktitle=skyblue!90!black,
  boxrule=0.6pt,
  width=0.95\textwidth
]

\textbf{Prompt 1} \\
You are given a movie context and a sequence of consecutive movie scenes, each composed of 5 images stacked vertically. 
Identify two scenes to answer the following question, which consists of two parts. 

\textit{Note: First, indicate which scene contains the answer to Part 1, and then specify which scene contains the answer to Part 2. 
Your answer format should be: 'Scene <number> for Part 1 and Scene <number> for Part 2.'}

\medskip
Context: \texttt{<CONTEXT>} \\
The question has two parts: \\
Part 1: \texttt{<PART1>} \\
Part 2: \texttt{<PART2>}

\medskip
\textbf{Output Samples:}  
Question: What advantage did Batman have when he initiated the brutal fight against Superman, and how was he able to gain such an advantage? \\
GT: Scene 7, Scene 3 \\
Answer: Scene 1 for Part 1 and Scene 2 for Part 2.

\medskip
Question: What event led to a memorial service being held in Metropolis, and how did such an event occur? \\
GT: Scene 5, Scene 2 \\
Answer: Scene 1 for Part 1 and Scene 2 for Part 2.

\medskip
\noindent\makebox[\linewidth]{\dotfill}
\medskip

\textbf{Prompt 2} \\
You are given a movie context and a sequence of consecutive movie scenes, each composed of 5 images stacked vertically. 
Identify two scenes to answer the following question, which consists of two parts.  

\textit{Note: First, indicate which scene contains the answer to Part 1, and then specify which scene contains the answer to Part 2. 
Your answer format should be: 'Scene 7 for Part 1 and Scene 2 for Part 2.'}

\medskip
\textbf{Output Samples:}  
Question: What advantage did Batman have when he initiated the brutal fight against Superman, and how was he able to gain such an advantage? \\
GT: Scene 7, Scene 3 \\
Answer: Scene 7 for Part 1 and Scene 2 for Part 2.

\medskip
Question: What event led to a memorial service being held in Metropolis, and how did such an event occur? \\
GT: Scene 5, Scene 2 \\
Answer: Scene 7 for Part 1 and Scene 2 for Part 2.

\end{tcolorbox}
\captionof{figure}{Various test prompts, but the model fails to generate reasonable answers.}

\newpage

\begin{tcolorbox}[
  colback=white,
  colframe=skyblue!80!black,
  title=Prompts and Outputs,
  coltitle=white,
  fonttitle=\bfseries,
  colbacktitle=skyblue!90!black,
  boxrule=0.6pt,
  width=0.95\textwidth
]

\textbf{Prompt 3} \\
You are given a movie context and a sequence of consecutive images, numbered starting from 1. 
Your task is to identify two images to answer the following questions. 

\textit{Note: First, indicate which image contains the answer to Part 1, and then specify which image contains the answer to Part 2. 
Your answer format should be: 'Image No.<number> for Q1 and Image No.<number> for Q2.'}

\medskip
\textbf{Output Samples:}  
Q: What advantage did Batman have when he initiated the brutal fight against Superman, and how was he able to gain such an advantage? \\
GT: Scene 7, Scene 3 \\
Answer: Image No. 1 for Q1 and Image No. 2 for Q2.

\medskip
\noindent\makebox[\linewidth]{\dotfill}
\medskip

\textbf{Prompt 4} \\
You are given a movie context and a sequence of consecutive images. 
Your task is to identify the image that best matches each question. 

\textit{Note: In your answer to each question, explain why you chose this image and specify which image it is in the sequence.} 

\medskip
\textbf{Output Sample (excerpt):}  
Q1: The image that best matches this question is Image No. 1. This image shows Batman holding a Kryptonite weapon, which is a significant advantage for him in the fight against Superman. \\
Q2: The image that best matches this question is Image No. 2. This image shows Batman wearing an armored exoskeleton, which allows him to withstand Superman's attacks.


\medskip
\noindent\makebox[\linewidth]{\dotfill}
\medskip

\textbf{Prompt 5} \\
You are given a movie context and a sequence of consecutive images. 
Your task is to find the answers to the following two questions by identifying the image that best matches each question. 

Note: In your answer, first explain why you chose that image, then count the number of this image in the sequence (e.g., ``This image is the 1st image in the sequence.").

\medskip
\textbf{Output Sample (excerpt):}  
Q1: The image that best matches this question is the 1st one in the sequence. This image shows Batman holding a Kryptonite weapon, which is a significant advantage for him in the fight against Superman. \\
Q2: The image that best matches this question is the 2nd one in the sequence. This image shows Batman wearing an armored exoskeleton, allowing him to withstand Superman's attacks and gain the upper hand.

\end{tcolorbox}
\vspace{-0.5em}
\captionof{figure}{Various test prompts, but the model fails to generate reasonable answers (continued).}\label{fig:different_prompt_2}

\section{Examples}
\begin{figure*}[h] 
    \centering
    \includegraphics[width=1\linewidth]{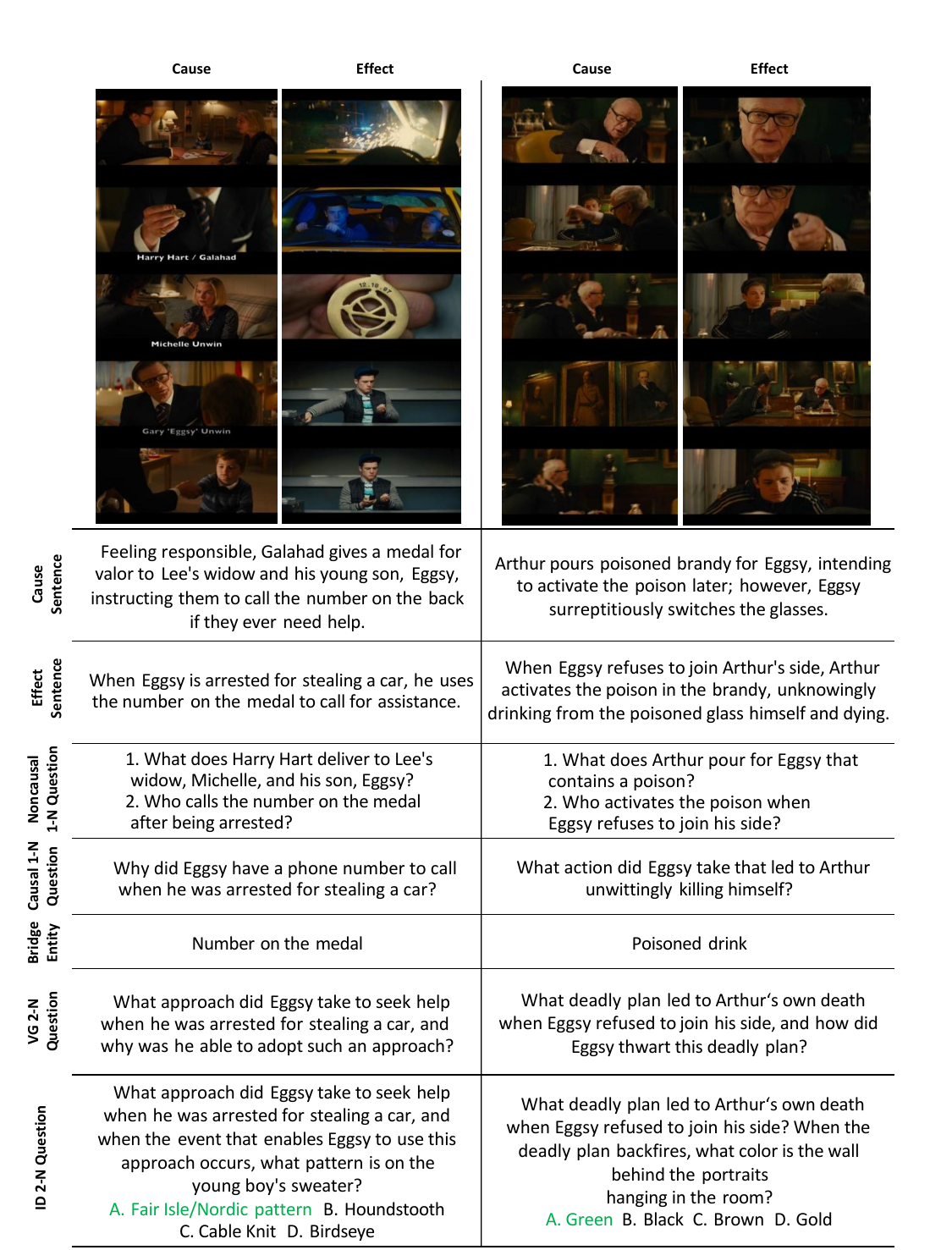} 
    \caption{The examples from our dataset.}
    \label{fig:examples}
\end{figure*}

\begin{figure*}[h] 
    \centering
    \includegraphics[width=1\linewidth]{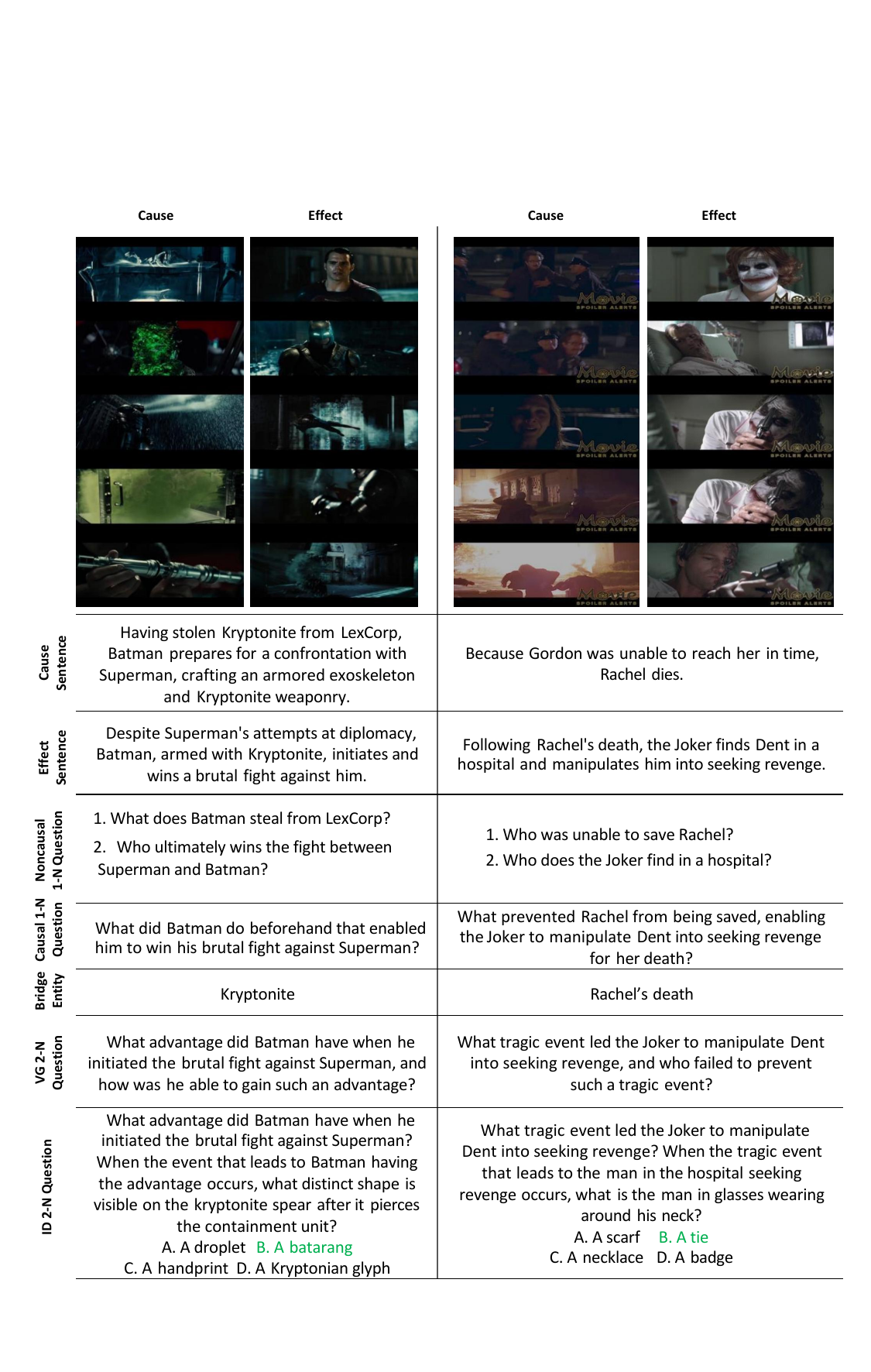} 
    \caption{The examples from our dataset (continued) .}
    \label{fig:examples_2}
\end{figure*}

We present some examples from the dataset in the Fig.~\ref{fig:examples} and Fig.~\ref{fig:examples_2}, covering all types of questions.

